\documentclass{article}

\usepackage[preprint]{neurips_2026}

\usepackage{microtype}
\usepackage{graphicx}
\usepackage{subcaption}
\usepackage{booktabs}
\usepackage{enumitem}

\usepackage[T1]{fontenc}
\usepackage[utf8]{inputenc}

\usepackage{hyperref}
\usepackage{xurl}
\usepackage{amsmath}
\usepackage{amssymb}
\usepackage{mathtools}
\usepackage{amsthm}
\usepackage{xcolor}

\usepackage[capitalize,noabbrev]{cleveref}

\theoremstyle{plain}

\theoremstyle{definition}

\theoremstyle{remark}

\usepackage{listings}
\usepackage{algorithm}
\usepackage{algorithmic}
\usepackage{tikz}
\usetikzlibrary{arrows.meta, positioning, calc, fit, backgrounds, shapes.geometric, shadows, matrix}

\title{Escaping the Cognitive Well: Efficient Competition Math with Off-the-Shelf Models}

\author{%
  Xingyu Dang\thanks{Equal contribution.} \\
  Princeton University \\
  \And
  Rohit Agarwal$^{*}$ \\
  Princeton University \\
  \And
  Rodrigo Porto \\
  Princeton University \\
  \And
  Anirudh Goyal \\
  \And
  Liam Fowl$^{*}$ \\
  Princeton Language and Intelligence \\
  \And
  Sanjeev Arora \\
  Princeton Language and Intelligence \\
  \texttt{arora@cs.princeton.edu} \\
}

\begin{document}

\maketitle

\begin{abstract}

 In the past year, custom and unreleased math reasoning models~\citep{deepmind2025imo, wei2025openaiimo} reached gold medal performance on the International Mathematical Olympiad (IMO). Similar performance was then reported using  large-scale inference on publicly available models~\citep{luong2025towards, shao2025deepseekmath} but at prohibitive costs (e.g., \$$3000$ per problem). In this work, we present an inference pipeline that attains best-in-class performance on IMO-style math problems at an average inference cost orders of magnitude below competing methods while using only general-purpose off-the-shelf models. Our method relies on insights about {\em grader failure} in solver-grader pipelines, which we call the \emph{Cognitive Well} (iterative refinement converging to a wrong solution that the solver as well as the pipeline's \emph{internal} grader consider to be basically correct). Our pipeline addresses these failure modes through \emph{conjecture extraction}, wherein candidate lemmas are isolated from generated solutions and independently verified alongside their negations in a fresh environment (context detachment). On IMO-ProofBench Advanced (PB-Adv), our pipeline achieves 87.6\% performance using Gemini 3.1 Pro with an average cost per question of $\sim$ \$9. This surpasses the performance of the DeepThink IMO gold-winning model, and more than triples the success rate of the next best publicly accessible pipeline operating at affordable cost\footnote{DeepSeekMath V2 reaches $61.9\%$ but at an estimated \$3000/question, making it inaccessible to most researchers. Among pipelines costing under \$300/question, the next best is \citet{huang2025winning} at $24\%$ expert-graded accuracy.}.

\end{abstract}

\section{Introduction}
Reasoning-based models have shown increasingly impressive progress on mathematical tasks. However, publicly available models cannot reliably solve difficult math problems during a \emph{single call} \citep{balunovic2025matharena}. Instead, they require a staged solution process, with a candidate solution passing through many rounds of grading, feedback, and refinement. In summer 2025, leading AI companies announced that their customized pipelines with models trained to do  ``parallel thinking'' had achieved gold-level performance at the 2025 International Math Olympiad (IMO) \citep{deepmind2025imo, wei2025openaiimo}. Soon after, other groups reported similar performance using  large-scale agentic pipelines on publicly-available models.
For example, \citet{huang2025winning} achieved gold medal performance via multiple parallel runs, each involving around $20$ rounds of interaction, with each round in turn involving a ``Grader'' critiquing the current proof, and the ``Solver'' refining the proof to address the feedback. In their setup, solving an IMO level math problem may use up to $100$ re-runs of the pipeline, which could amount to \textit{thousands} of LLM calls per problem \citep{luong2025towards}!
A similar pipeline based on a  specialist model by DeepSeek \citep{shao2025deepseekmath} reached gold-level performance by allowing up to $64$K LLM calls for a maximum budget of $8$ billion tokens per problem, which would convert to an estimated $3000$ US dollars in API costs. Such pipelines  remain too costly for most researchers, let alone enthusiasts. 

\begin{figure*}[th!]
    \centering
    \includegraphics[width=0.8\textwidth]{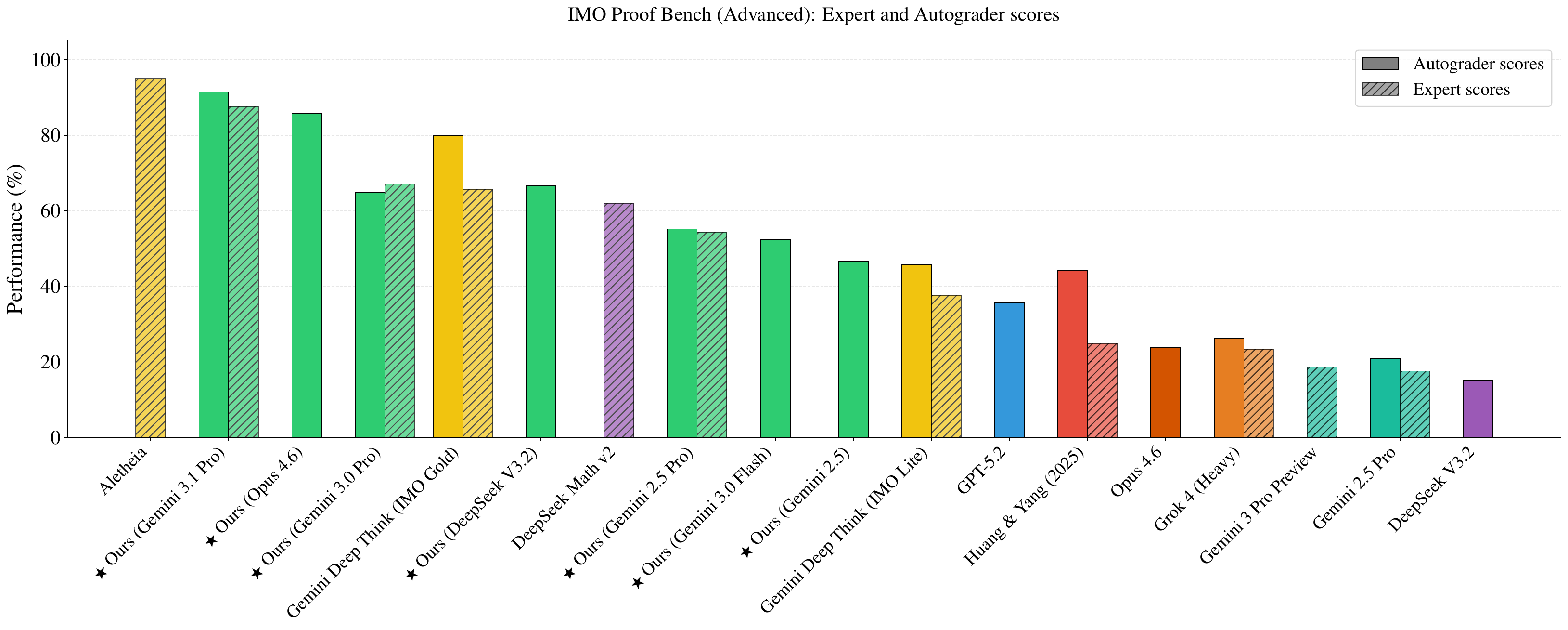}
    \caption{Results for our proposed method versus competing methods and models. Both expert graded and autograded results included (where possible). Our proposed method outperforms all open-source pipelines, while being significantly cheaper than these pipelines (cf. Figure \ref{fig:pareto}) and ranks second only to a custom and unreleased pipeline. Comparison results taken from \citet{luong2025towards} where applicable, and from \citet{shao2025deepseekmath}. Autograder results calculated using the resources published in \citet{luong2025towards}.}
    \label{fig:autograder_performance}
\end{figure*}

In an effort to measure the rapid progress in this area, a recent paper \citep{luong2025towards} introduces evaluations (and accompanying leaderboards) to measure LLM-based systems' abilities to solve/grade IMO style problems. Of special note is IMO-ProofBench (Advanced) --henceforth called PB-Adv-- consisting of $30$ curated IMO-style questions that  challenge today's models. Proprietary math models Google Deep Think (Lite) and Deep Think (Gold) score  $39\%$ and $65\%$ respectively but their compute usage  is unreleased. A concurrent work to ours, Aletheia \citep{Aletheia}, builds off the newest (unreleased and custom) Gemini Deep Think and achieves state-of-the art performance on PB-Adv, with unspecified cost and compute. Another competing pipeline, DeepSeekMath V2, reaches $61.9\%$ but, as noted, allows up to $3000$ USD/question in  API cost. The \citet{huang2025winning} pipeline, running on Gemini 2.5 pro with a budget exceeding $300$ USD/question, attains only $24$\% accuracy. Smaller, custom trained models and pipelines offer more affordable options, but do not achieve competitive performance on the more challenging PB-Adv set \citep{qednano2026}. We introduce a new pipeline that achieves best-in-class performance on a budget that is more than an order of magnitude smaller than comparable methods (cf. Figure~\ref{fig:pareto}).

\subsection{Identifying Failure Modes}
\label{subsec:failure_modes}

Why do the aforementioned pipelines need expensive parallelism for hard problems? It helps mitigate two phenomena:

\noindent{\bf Cognitive Plateau:} The solver has made some progress, but much of the problem remains unsolved. On successive proofs, the grader points out flaws, but the solver struggles to fix them all, and the grade does not meaningfully change. We call this a {\em cognitive plateau}. Massive parallelism acts as a form of exploration and discovers rare good ideas, which the grader recognizes as progress. 

\noindent{\bf Cognitive Well:} Iterative refinement converges to an imperfect (even dead wrong) solution that has high enough logical  consistency that the grader --usually sharing the solver's parameters and sometimes its context window--scores it positively, overlooking the fatal flaws to be  ``minor slips.''  For example, \citet{luong2025towards} reports that on $28$ out of $30$ PB-Adv problems, the full \citet{huang2025winning} pipeline produced a solution that received a perfect $7/7$ from the pipeline's internal grader in $5$ independent calls, but a majority of times the solution was completely wrong--as judged by the benchmark's grader provided with a reference  human solution\footnote{Our investigation revealed that this arises from the solver exploiting a logical loophole in the grader prompt. This is suggestive of a form of reward hacking.}. We conjecture that such failure modes are the reason \citet{shao2025deepseekmath} grades each answer $64$x times in parallel. We give a more in-depth discussion of the \citet{huang2025winning} grader in Appendix~\ref{ap:grader_analysis}.

\begin{figure*}[ht]
    \centering
    \includegraphics[width=0.75\textwidth]{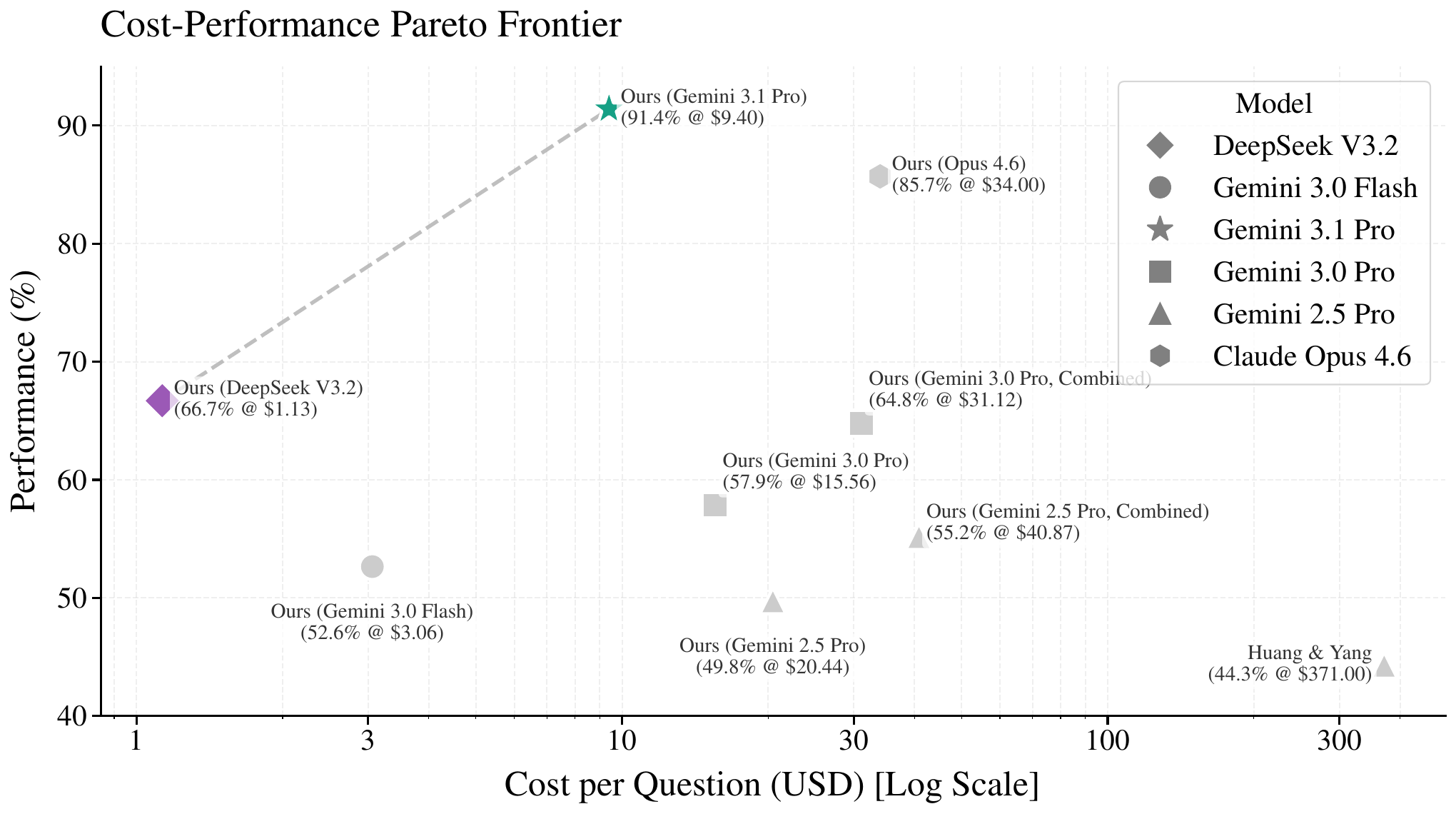}
    \caption{Pareto frontier of cost vs. performance measured using the autograder in \citet{luong2025towards}.  ``Single" denotes that the output of a \emph{single} run of the pipeline. ``Combined" denotes our standard algorithm for PB-Adv which uses a further model call to choose from two parallel pipeline runs - cf. Section \ref{subsec:pipeline_arch} ``Post Enhancement and Aggregation"). All costs calculated with January 2026 pricing.}
    \label{fig:pareto}
\end{figure*}

Appendix \ref{ap:cognitive_well_analysis} presents examples that illustrate how our method circumvents the above failure modes.  
The improvements demonstrated in Figures \ref{fig:autograder_performance} and \ref{fig:pareto} stem from new conceptual fixes we propose for  these issues, as described in Section~\ref{sec:methods}. 
The importance of the grader role is emphasized in~\citet{luong2025towards} by their GraderBench evaluation and we report experiments on this evaluation in \cref{subsec:gradereval}.

\subsection{Misconception: ``Grader as compass''} 
\label{subsec:gradermyth}
It may appear fraught to allow the same underlying model to produce proofs as well as grade them (via  two copies running solver and grader processes). 
But the intuition supporting this practice is that grading a math proof\footnote{That is, checking that each line follows from previous lines and the final step contains the desired conclusion.}  seems easier than {\bf finding}  a correct proof\footnote{The intuition is formalized as the famous $P \neq NP$ conjecture \cite{arora2009computational}.},  so one expects any given model to be  \emph{better} as grader than it  is as solver. Thus {\em in principle} a model should be able to reliably grade (via an independent call with clean context) its own solution.  

However, the above intuition only supports {\bf binary} (correct/incorrect) verification. In practice it is tempting to  use grader feedback as a {\em measure of progress}. This is untrustworthy as the grader may not know how to solve the question!  The question may even be functionally impossible to solve in a single model call. Thus, upon detecting  gaps in a proof, the grader cannot discern  how serious/fixable they are, and thus is unable to  estimate the point penalty. Grader scores can fluctuate in counterintuitive ways: a new idea may represent progress for the solver, but may make the grader note new hurdles and reduce its score.
The {\em Cognitive Plateau} arises due to this unreliable signal.
 The {\em Cognitive  Well} is more counter-intuitive: the grader's near-total acceptance of a deeply flawed solution. We hypothesize that the cause is the multi-step interactive process where the solver improved the proof in response to the grader's feedback. The final flawed proof thus looks largely or completely OK to the computationally-limited grader ---this is reminiscent of {\em reward hacking} and also of stable local minimum in optimization theory. Motivated by this finding, our method limits the grader's feedback to a short list of mistakes, some phrased as questions.

\noindent{\bf Related Area: Formal Proofs.}
A parallel research pillar in AI for Math involves models that generate formally verifiable proofs  --usually in LEAN \citep{moura2021lean}--instead of in natural language; see \citet{GoogleDeepMind2024IMO,lin2025goedel, lin2025goedelv2, ren2025deepseek}. For this work, we focus on natural language proving as this represents the forefront for mathematical abilities of current off-the-shelf LLMs. 

\subsection{Contributions and Paper Organization}

The following are our core contributions. 
\begin{enumerate}
    \item A  powerful new solution pipeline for competition math problems using only off‑the‑shelf models. It achieves best-in-class performance and costs up to two orders of magnitude lower than competing methods. To achieve this, we describe new solving methodologies, as well as uncovering insights into current pipelines' failures.
    \item Identifying the “Cognitive Wells” trap in solver–grader systems and providing an effective solution for it involving a conjecture-based context-detachment mechanism (prove conjecture and its negation in fresh context) that measurably reduces false-positive wells and boosts performance (cf. Section \ref{subsec:pipeline}).
    \item New methods for designing more effective graders for use within a math-solving pipeline, as well as new suggestions for how to evaluate graders (cf. \cref{subsec:gradereval}).
    \item Open-sourced prompts and methodology ---ready to run on public models. We show the efficacy of our pipeline on a range of frontier and open-weight models---Gemini 3 Pro/Flash, Gemini 2.5 Pro, Claude 4.6 Opus, DeepSeek-V3.2---at an affordable cost; this significantly lowers the barrier to entry for enthusiasts and researchers.

\end{enumerate}

\section{Method Overview}
\label{sec:methods}

Solver-grader pipelines usually involve an {\bf orchestrator} process  that dynamically extracts contexts from previous LLM calls and uses them to prompt subsequent calls. We created a {\em ``dialectic based''}  prompt-engineering methodology whereby, instead of asking the model to act as, say, an IMO grader,  we instruct it  to use all provided materials to conduct a multi-round dialectic among some named personas following certain rules.  For example, one persona may be extremely picky about slips in the proof, whereas another might be asked to  argue that it represents standard practice.  Tweaks to  each rule changes the final dialectic in fairly predictable ways (see Appendix~\ref{sec:prompts}) and adding new personas allows new functionalities (e.g., conjecture extraction). In a single model call our dialectic-based prompts seem to perform somewhat better than  more conventional prompts (see Appendix~\ref{ap:solver_ablation} and Table~\ref{tab:solver_comparison_appendix}), but overall, we stress the main benefit is convenience and precise control\footnote{We refer to our pipeline as ``Momus pipeline,''  with Momus being  a picky persona that was named after a mythical figure in Greek philosophy.}.

\subsection{Pipeline Design Ideas}
\label{subsec:pipeline}

To address the failure modes described in Section \ref{subsec:failure_modes} while maintaining computational efficiency, our pipeline rests on three core design principles:

\begin{enumerate}
    \item \textbf{Narrow Width (Limited Parallelism):} 
    Brute-force parallelism (e.g., maintaining 64 parallel branches) incurs high costs and makes monitoring difficult. Inspired by recent  PDR method \citep{madaan2025rethinking}, we limit parallelism to a modest level (typically $K=4$ concurrent solutions). The next group of solutions is generated conditioned on the current state, which is the previous group of solutions, feedback from the grader, as well as the results of the conjecturing process described below. 

    \item \textbf{Conjecture Extraction:} 
    Pipelines often stall because the internal grader cannot distinguish between a ``minor slip" and a fatal gap. To resolve this, we employ a \textit{Solver with Conjectures} prompt. This module is given the current set of stalled proofs and asked to explicitly formalize logical gaps as self-contained conjectures (and also formalize their negations). This forces the model to define exactly what is missing to complete a proof, transforming vague uncertainty into falsifiable mathematical statements.

    \item \textbf{Contextual Detachment:} 
    The ``Cognitive Well" occurs when a grader becomes ``contaminated" by the context of a plausible-but-wrong proof.  Combating this requires stripping extracted conjectures from their parent context, and feeding them --as well as their negations--as new problems to fresh solver instances. If that results in proving the \textit{negation} of a necessary conjecture, including this as ``additional material" for the solver in the next step can nudge it to discredit the ideas of the stalled proof, and think anew.
\end{enumerate}

\subsection{Pipeline Architecture}
\label{subsec:pipeline_arch}

Our inference pipeline organizes these principles into a stateful loop. In previous pipelines the state is a single candidate solution --or a few solutions. Our conjecture process turns the state into a set of candidate proofs for the original problem, grader feedback,  as well as proofs/disproofs for intermediate conjectures. The process has three phases:

\textbf{Phase I: Limited Exploration} \\
We generate a small set of candidate solutions (typically $K=4$) using the Dialectic Solver.
\begin{enumerate}
    \item \textbf{Generation:} The model produces candidate proofs $P_1 \dots P_K$.
    \item \textbf{Adversarial Grading:} A separate instance of the model grades each $P_i$. The grader is prompted to use a ``guilty until proven innocent" framework for listing specific logical gaps.
    \item \textbf{Outcome:} If a solution receives a perfect score ($7/7$) across three independent grading runs, it is accepted. If not, the pipeline enters Phase II.
\end{enumerate}

\textbf{Phase II: Contextual Detachment} \\
This phase targets the \emph{Cognitive Plateau} and \emph{Cognitive Well}. When solutions are plausible but unverified (or stuck at a score $\leq 6/7$), we assume the verifier is ``contaminated" by the context of the incorrect proof and execute the following:
\begin{enumerate}
    \item \textbf{Extraction:} We prompt the model to identify load-bearing conjectures $\{C_1, \dots, C_m\}$ from top-scored candidate proofs. These are statements which, if true, would help complete the proof.
    \item \textbf{Isolation:} Each conjecture $C_j$ is stripped of its parent proof context as an independent problem. 
    \item \textbf{Bisection:} For each conjecture, we instantiate two independent copies of the Phase I procedure in parallel to prove both $C_j$ and its negation $\neg C_j$.
    \item \textbf{Resolution:} If one of $C_j, \neg C_j$ is proven, it is added to a persistent global memory ($\mathcal{M}_{lemma}$) as a proven lemma. (conjecture or negation) 
\end{enumerate}

\textbf{Phase III: Refinement with Global Memory} \\
The pipeline launches a new round of solvers. However, all but one of these solvers are no longer zero-shot. They are injected with the ``Global Memory" $\mathcal{M}_{lemma}$ containing the independently verified lemmas from Phase II. If a positive conjecture was proven, this allows the new solver to make further progress on the suggested proof ideas. If a negative conjecture was proven, this allows us to potentially escape a Cognitive Plateau and change ideas. We also provide any partial progress on proving conjectures in the previous round, as this may be helpful to a solver. The one zero-shot solver remains to keep diversity and avoid falling into a Cognitive Well.

\textbf{Post Enhancement and Aggregation.}
If the conjecture iteration budget is exhausted without yielding a verified solution, the pipeline executes a final post-enhancement step. Here, we run 2 parallel sessions of Phase II, where the conjecture extractor is rewriting the proof based on the conjectures. If any of the 2 parallel session's conjectures have been fully proved, we include the corresponding written proof in $\mathcal{M}_{lemma}$ and launch a session of Phase III for the final solving. (Algorithm~\ref{alg:main_pipeline} Phase 4 presents a simplified version of this step: it runs a single gap-extraction and verification pass followed by one final \textsc{DialecticSolve} call.)

To maximize performance, we may run two instances of the entire pipeline in parallel. If so, a final ``Judge" combiner persona compares the final outputs of both pipelines. We employ this strategy for more difficult questions, but opt for a more efficient single pipeline run for simpler problems (cf. Figure \ref{fig:pb_basic}).

For a more formal description of our algorithm, please refer to Appendix \cref{ap:alg_details}, and specifically Algorithm \ref{alg:main_pipeline}. Our pipeline is generally model agnostic - although performance does scale with base model capability. We provide ablations for important parts of our pipeline in Section \ref{sec:results} and in the Appendix.

We also provide high-level visualization of this pipeline architecture in Appendix Figure \ref{fig:v19-pipeline}.

\section{Results}
\label{sec:results}
IMO-Bench represents the most natural domain in which to test our methods. While the difficult subset PB-Adv comprises only $30$ questions, it is  designed to be difficult for current systems and  relatively contamination-free.

IMO ProofBench judges correctness using the full proof/reasoning, rather than just a numeric final answer/expression.  
It comes with an  autograder prompt provided with detailed gold solutions and scoring rubrics ~\citet{luong2025towards}. Its ratings correlate strongly with human judgments ($r = 0.93$; see Section~\ref{subsec:gradereval}) but can be less accurate on AI generations. Expert human grading is thus the gold standard. 

Because of limited resources and the need to compare to results previously reported, our headline numbers come from four Gemini models: Gemini 3.1 Pro, Gemini 3.0 Pro, Gemini 3.0 Flash, and Gemini 2.5 Pro. However, we additionally report results with further backbones (Claude 4.6 Opus, the open-weight DeepSeek-V3.2) in Section~\ref{subsec:cross_backbone}.

 While most of our reported numbers correspond to the autograder, 
 we also used human graders (IMO medalists, math competition problem writers, and mathematicians) for one full run  each of Gemini 2.5 pro and Gemini 3 pro. We found high consistency between these (blind) expert scores and autograder scores, which suggests our pipeline may avoid discrepancies between the two grading methods present in other works (e.g. see Figure~\ref{fig:autograder_performance} for the performance of DeepThink models and  HY pipeline as reported in \citet{luong2025towards}).  We think that  divergence between the Autograder and human graders should be evaluated as possible indications of in-context reward hacking mentioned in Section~\ref{subsec:failure_modes}.

In Figure \ref{fig:autograder_performance} we present a comparison of our method's performance on IMO Proof Bench (Advanced) to a wide range of other frontier models and pipelines using both autograder performance and expert grader performance (where possible). According to the expert human grading, our pipeline scores second - only to the concurrent (and unreleased) Aletheia model from Google DeepMind that was announced in February 2026. Notably, our best run outperforms the earlier Deep Think IMO-Gold model, and the massively parallel (and expensive) pipeline incorporating the specialst DeepSeek Math v2 model \citep{shao2025deepseekmath}. Also of particular note is our performance compared to the \citet{huang2025winning} pipeline where, when using the same exact base model, we more than double the performance (according to expert grading) while operating at a budget an order-of-magnitude lower.

\subsection{Generalization Across Backbones}\label{subsec:cross_backbone}

The Gemini family powers both the leading academic pipeline of \citet{huang2025winning} and the 2025 IMO-gold winning DeepThink system, which is why our headline numbers use it. To verify that the gains are not specific to Gemini, we additionally ran the pipeline with Claude 4.6 Opus, the open-weight DeepSeek-V3.2 as the backbone for all roles (solver, grader, and conjecture-extractor). Appendix Table~\ref{tab:cross_backbone} reports the autograded average score on PB-Advanced for each backbone alongside the score of a single base call to the same model.

Across all three additional backbones our pipeline drives substantial gains. The largest is on Claude 4.6 Opus, which jumps from a $23.8\%$ base call to $85.7\%$; the open-weight DeepSeek-V3.2 jumps from $15.2\%$ to $66.7\%$. These results indicate that the gains attributed to conjecture extraction and contextual detachment are not artefacts of a single model family.

\subsection{Performance on Additional Benchmarks}\label{subsec:more_benchmarks}

PB-Adv is the gold standard for difficult competition level mathematics (for example, this benchmark serves as the main barometer for Aletheia, and DeepSeek Math v2). However, PB-Adv comprises only $30$ questions, and therefore is a relatively narrow basis for evaluation. Thus, we additionally ran our pipeline (with Gemini 3.0 Flash, our cheapest backbone) on two larger benchmarks: \textbf{ProofBench} ($145$ USAMO/IMO/Putnam questions, distinct from IMO-ProofBench) and \textbf{IMO-AnswerBench} ($400$ questions, final-answer evaluation). Baselines for both benchmarks are taken from QED-Nano \citep{qednano2026}. Appendix Table~\ref{tab:more_benchmarks} reports the results. Our pipeline outperforms all baselines on both benchmarks despite using our weakest backbone. Note that broader evaluations remain cost-prohibitive at academic budgets: running our pipeline on the $400$ IMO-AnswerBench problems with Gemini 3.0 Flash alone costs roughly \$$1{,}000$.

\subsection{Solver Prompt Evaluation}\label{results:solver}

\label{sec:results_solver}

To isolate the source of our performance gains, we examined whether our internal dialectic (Momus) prompting strategy provides an inherent advantage over standard prompting. We conduct a controlled comparison against the solver prompt from \citet{huang2025winning} on PB-Basic  (30 problems, 4 independent runs each). 

We observed a modest performance difference between the two strategies: the Momus solver achieved an average-of-4 score of 5.10/7, while the Huang-Yang solver got 4.92/7. Detailed results are available in Appendix \ref{ap:solver_ablation}. Overall, the dialectic approach does not seem to yield a significantly better solver. 
This parity suggests that the significant performance difference observed in Figure \ref{fig:autograder_performance} stems primarily from our pipeline's other innovations—specifically, grading, conjecture extraction and contextual detachment—rather than the initial solver prompt formulation. Our dialectic approach was used for prompt design for these other roles as well. 

\subsection{Grader Evaluation}\label{subsec:gradereval}
 Because the internal grader in a pipeline is of paramount importance for the success of the solver, we additionally conduct evaluations on the Graderbench benchmark in \citet{luong2025towards}. Graderbench evaluates \emph{grader performance} using a dataset of AI-generated solutions to PB-Adv questions with accompanying human grades on $0$-$7$ scale.  {\bf Accuracy} is measured via a $4$-way classification task with labels {\bf $0/1/6/7$}:  label {\bf $1$} standing for $\{1, 2, 3\}$ and {\bf $6$} for $\{4, 5, 6\}$, since that aligns with human grading practices.
 {\bf }
 In this section we use it to evaluate our grader and also to shed light on ways in which the benchmark may need rethinking. 

 In Section~\ref{subsec:gradermyth}, we cautioned against taking    the internal grader's numerical score seriously, since accurate  scoring  requires estimating  what fraction of the problem is unsolved. This is infeasible given the grader's compute limitations, and no access to a reference solution/rubric. Thus our pipeline mostly ignores  the numerical grade except for tie-breaking and exiting early due to reaching a perfect score (with zero slips found in the proof) thrice in a row. Our {\bf Momus grader}  is designed with Cognitive Wells in mind: it prioritizes {\bf strictness}. 

 It is still informative to see evaluations on GraderBench for:
  (i) Our vanilla \textbf{Momus Grader} that is used in our solution pipeline, (ii) a variant \textbf{Momus Grader w/ $3$ Pipeline Solutions} inspired by the above-mentioned argument of Section~\ref{subsec:gradermyth}. To give  our Momus grader  the benefit of more compute, we provide it with three sample solutions (from evenly spaced runs) generated during our pipeline's attempt to solve the same problem. Even though our pipeline fails to solve many problems, one hopes that the intermediate attempts nevertheless help the grader understand the problem's intricacies.

GraderBench evaluation centers on (i) \textbf{Accuracy (Acc)} on the classification task, and (ii)   
\textbf{Mean Absolute Error (MAE)}: As in \citet{luong2025towards}, model-predicted scores are rounded to $0/1/6/7$, and compared with human grades on a 7-point scale. 

 All our experiments in this section used Gemini 2.5 Pro.  Numbers in the table for other models are taken from \citet{luong2025towards}  which in turn evaluated using the \textbf{ProofAutoGrader} prompt without the reference human solution. 
 
\begin{table}[h]
\centering
\caption{Comparison of Momus Prompts vs. Other models on GraderBench (1000 datapoints). No reference solution provided. Non-Momus models are run with the ProofAutoGrader prompt. MAE is expressed as a percentage of the maximum score of 7.}
\label{tab:other_models_grader_metrics}
\begin{tabular}{lcc}
\toprule
\textbf{Method} & \textbf{Acc $\uparrow$} & \textbf{MAE $\downarrow$} \\
\midrule
 Gemini 2.5 Pro & 44.3\% & 30.2\% \\
\textbf{Momus} & 28.1\% & 25.9\% \\
 o4-mini (high reasoning) & 47.3\% & 25.2\% \\
  Gemini 2.5 Deep Think & 52.5\% & 20.5\% \\
  o3 & 54.0\% & 20.2\%  \\
\textbf{Momus w/ p. sols} & 49.8\% & 19.9\% \\
  Gemini 2.5 Deep Think (IMO Gold) & 50.2\% & 18.4\%\\
\bottomrule
\end{tabular}
\end{table}

\cref{tab:other_models_grader_metrics} shows that Momus with pipeline solutions is competitive with the best graders.

\noindent{\bf Does GraderBench performance capture usefulness?} 

The short answer: No! In Table~\ref{tab:other_models_grader_metrics} the vanilla Momus grader looks worse than the Momus grader with pipeline solutions yet provides better performance in our  solution pipeline. Understanding reasons for this helps highlight some subtleties ignored in GraderBench.

Pipeline performance (especially when using low parallelism) depends closely on the following metrics, with {\em positive} considered a score of $6$ or higher; a {\em negative} considered a score of $5$ or lower\footnote{If we combine the 0 and 1-3 buckets, then the classification accuracies in Table~\ref{tab:grader_metrics} become more comparable. ProofAutoGrader is 73.5\% and Momus with pipeline solutions is 72.0\%, while base Momus gets 53.7\%.}.
\begin{itemize} 
\item \textbf{False Positive Rate (FPR)}: $P(\text{pred}\ge 6 \mid \text{human}\le 5)$. In a pipeline setting this is a critical failure mode, a form of ``Cognitive Well'' with increased risk of submitting an incorrect solution.

\item \textbf{False Negative Rate (FNR)}: $P(\text{pred} \le 5 \mid \text{human} \ge 6)$. Grader being over-strict  is not a critical failure; it triggers another round of  exploration that often improves the solution.
\end{itemize}
The above metrics are shown for  a simplified presentation in Table~\ref{tab:grader_metrics}; the full picture involves confusion matrices in  \cref{fig:grader_confusion}.

\begin{table}[h]
\centering
\caption{Comparison of Momus Prompts vs.\ ProofAutoGrader (with reference solution) on GraderBench (1000 datapoints). MAE is expressed as a percentage of the maximum score of 7.}
\label{tab:grader_metrics}
\begin{tabular}{lcccc}
\toprule
\textbf{Method} & \textbf{FPR $\downarrow$} & \textbf{FNR $\downarrow$} & \textbf{Acc $\uparrow$} & \textbf{MAE $\downarrow$} \\
\midrule
 \textbf{Momus} & 17.2\% &26.1\%	& 28.1\% & 25.9\% \\
ProofAutoGrader & 31.7\% & 6.2\%  & 59.4\% &  17.6\% \\
\textbf{Momus w/ p.sols} & 26.9\% & 8.8\% & 49.8\% & 19.9\% \\
\bottomrule
\end{tabular}
\end{table}

 Momus grader's lower FPR rate reflects an emphasis on a safety margin: being picky helps mitigate Cognitive Wells. Pickiness increases FNR, but that is acceptable because the pipeline always ends by returning the best solution it found.  In fact, many solutions considered imperfect by Momus receive higher scores from the ProofAutoGrader.  

 This raises the crucial point: even the gold ProofAutoGrader might  not be best for use in a solution pipeline.   It prioritizes matching human grades on the benchmark which rarely fall between $2$ to $5$.  Its higher grading accuracy comes at the expense of worse FPR, which would be expected to degrade the accuracy if used to terminate a solution pipeline. A good middle ground is the {\bf Momus grader with pipeline solutions} which recovers much of the accuracy difference and almost matches the MAE for ProofAutoGrader. However, its FPR is somewhat higher than vanilla Momus, and experimentally this  appears to already \emph{hurt} its utility for the pipeline. In summary, a grader's value as a standalone scorer (e.g., as measured by GraderBench) may not be aligned with its role as a source of feedback in a larger pipeline. This tension—and the need for better metrics to capture it—remains an important area for future study.

\subsection{Conjecture Removal: Analysis}\label{subsec:conj_removal}

In order to investigate the added value of conjecture extraction, we isolate problems where the Gemini 2.5 Pro version of the pipeline is successful \emph{and} proved at least one conjecture. We then try omitting the conjecture stage entirely. To perform this ablation, we modify our method so that our pipeline simply loops over the parallel solving stage with the standard $4$ solvers and the usual Momus grader. Furthermore, we match token counts: that is, if our pipeline uses $T$ tokens, we allow the modified pipeline to also use up to $T$ tokens (early termination is allowed). We then select $5$ evenly spaced checkpoints throughout this pipeline's run and grade them with the ProofAutoGrader as well as the internal pipeline grader.
\begin{figure}[h!]
    \centering
    \includegraphics[width=0.65\linewidth]{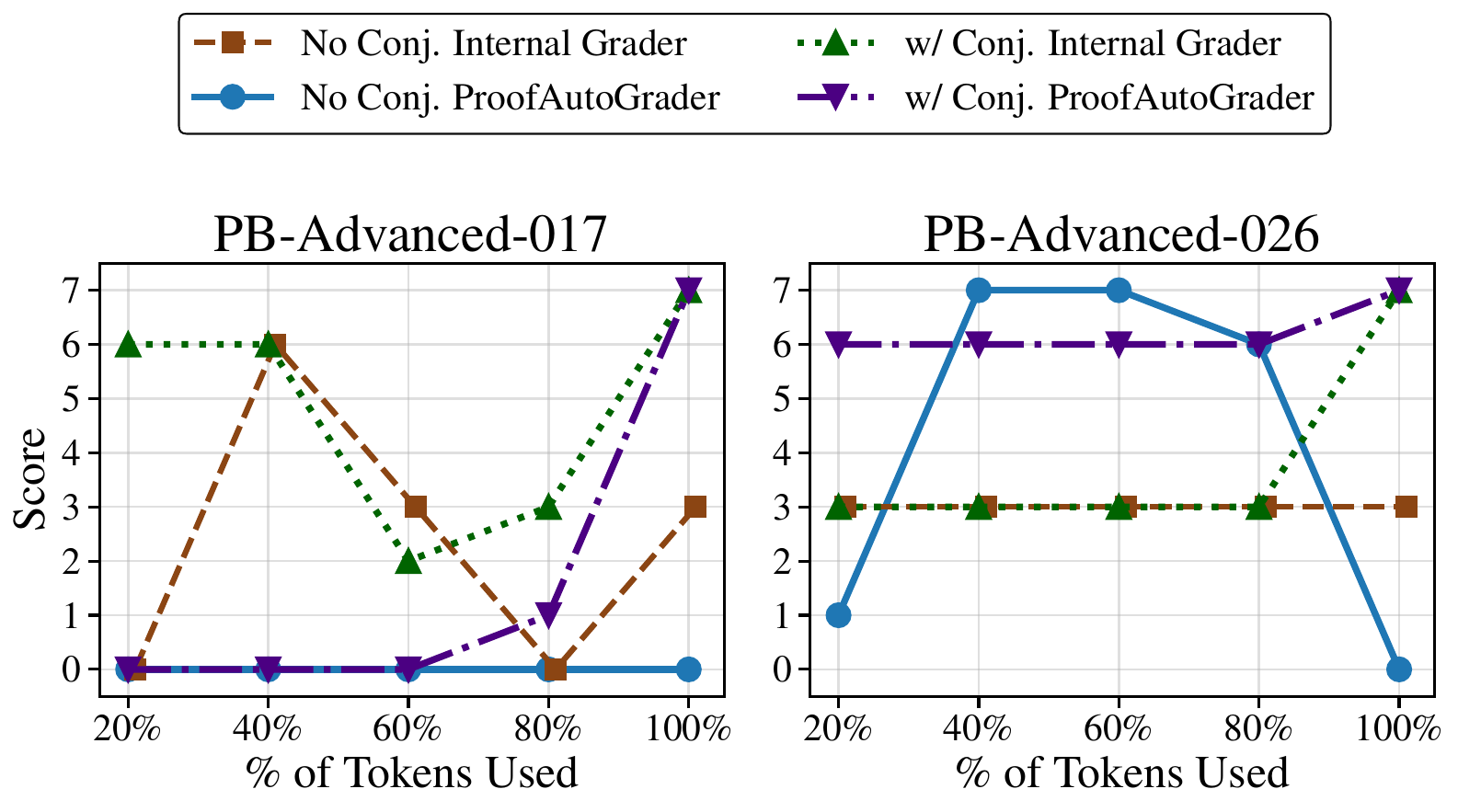}
    \caption{
    Case study of benefits of conjecture extraction/resolution on the solution pipeline. We run (a) the pipeline without any conjecture steps and (b) with conjecture steps. For both runs, we equalize token budgets. We sample
    solutions for five phases throughout the runs. We then grade these runs using both the pipeline's internal grader and the external ProofAutoGrader.
    We see that with conjectures, the ProofAutoGrader scores monotonically increase, while those without may jump wildly or flatline.  Spikes in conjecture-based solving correspond to proof/disproof of conjectures.}
    \label{fig:phase1_focused_traj}
\end{figure}

\cref{fig:phase1_focused_traj} shows the results on two representative problems. (Appendix \ref{ap:conj_removal_details} has more examples.) In particular on PB-Adv-017, we see that neither setup makes progress with respect to ProofAutoGrader grades for the first 60\% of the run. In the last 40\%,  proven conjectures accumulate and allow the model to score a perfect 7/7. Contrast this with  progress after removal of proved conjectures--the pipeline grader awards 6/7 to a solution  that is completely incorrect. This gets it stuck in the ``Cognitive Well,'' unable to make any better solutions. Appendix \ref{ap:conj_removal_details}, includes results on a broader class of problems, exhibiting how  inclusion of conjecture extraction reliably stabilizes and increases final  scores.

\paragraph{Aggregate effect at matched compute.} The per-problem trajectories above leave open whether the conjecture-removed pipeline could simply ``catch up'' given enough additional sampling. To rule this out, we ran a token-matched ablation on Gemini 2.5 Pro across all of PB-Advanced. We removed the conjecture-extraction stage entirely and scaled up the standard solver--verifier loop until each problem received up to $\sim\!2.1$M compute tokens---the per-problem budget our full pipeline uses on its hardest PB-Advanced problem. Even with this generous budget, the no-conjecture variant scored only $34.3\%$ on PB-Advanced, compared to $55\%$ for the full pipeline on the same backbone. The gap confirms that the gains we attribute to conjecture extraction and contextual detachment are not just an artefact of additional solver diversity or sampling budget.

\section{Conclusion}
Our work suggests that  reliable LLM theorem proving may not require massive parallelism and thinking budgets. Our efficient inference pipeline enables off-the-shelf models to reach similar performance at a fraction of the cost of comparable published methods. Central to our contribution is the identification of \emph{Cognitive Plateaus} and \emph{Cognitive Wells} wherein pipelines can become trapped in apparently plausible but ultimately incorrect reasoning paths that fool their own internal verifiers. We suspect such failure modes are deep-rooted in current solver/grader pipelines, and may extend to broader AI-for-Science systems. This phenomenon requires careful attention, and further exploration is left for future work, as is better grader evaluation.

We validate our pipeline on the recently released suite of Proof-Bench evaluations. On IMO Proof Bench-Advanced  we scored (at the time of the experiments) ahead of all but one competing methods, both released and unreleased, with performance verified by expert graders. Additionally, we probe the role of graders as standalone scorers versus feedback providers and analyze where the two roles may not align.

\begin{ack}
We thank IMO Medalists and Math Enthusiasts Minjae Kwon, Xinmiao Han, Liangjun Zhong, Geoffrey Sangston and Kaiyue Wen for their help human-grading many of the model-generated proofs. We acknowledge API access from Google Academic Project award.
This work was funded in part by ONR, Darpa AIQ (AI-Quantified), Schmidt Foundation, and Google DeepMind.
\end{ack}

\bibliographystyle{plainnat}
\bibliography{example_paper}

\clearpage
\newpage
\appendix
\section{Algorithm Details}
Here we provide a high-level schematic of our pipeline, and detailed algorithms.

\begin{figure}[ht!]
    \centering
    \scalebox{0.7}{
%

\definecolor{cGen}{RGB}{52, 152, 219}   
\definecolor{cGrade}{RGB}{142, 68, 173} 
\definecolor{cDist}{RGB}{136, 14, 79}   
\definecolor{cRes}{RGB}{241, 196, 15}   
\definecolor{cOut}{RGB}{39, 174, 96}    
\definecolor{cWarn}{RGB}{230, 126, 34}  
\definecolor{cGray}{RGB}{127, 140, 141} 

\tikzset{
    stagebox/.style={
        rectangle, rounded corners=10pt,
        draw=#1!70, line width=2pt, fill=#1!6,
        inner sep=8pt 
    },
    stagelabel/.style={
        font=\LARGE\itshape,
        text=#1!90,
        fill=white,
        rounded corners=4pt,
        draw=#1!25, line width=0.8pt,
        inner xsep=6pt, inner ysep=3pt
    }
}

\resizebox{\textwidth}{!}{%
\begin{tikzpicture}[
    font=\sffamily,
    basebox/.style={
        rectangle, rounded corners=6pt, align=center,
        draw=black!35, line width=2pt, fill=white,
        drop shadow={opacity=0.15, shadow xshift=1pt, shadow yshift=-1pt},
        inner xsep=10pt, inner ysep=8pt, 
        font=\LARGE
    },
    solver/.style={
        basebox, draw=cGen, fill=cGen!9,
        minimum width=24mm, minimum height=12mm
    },
    grader/.style={
        basebox, draw=cGrade, fill=cGrade!9,
        minimum width=24mm, minimum height=12mm
    },
    proc/.style={
        basebox, draw=cDist, fill=cDist!13,
        minimum width=48mm, minimum height=14mm
    },
    resbox/.style={
        basebox, draw=cRes, fill=cRes!18,
        minimum width=68mm, minimum height=16mm
    },
    decidebox/.style={
        diamond, aspect=2.2, align=center,
        draw=cWarn, fill=cWarn!18, line width=2pt,
        inner sep=6pt, font=\LARGE,
        drop shadow={opacity=0.15, shadow xshift=1pt, shadow yshift=-1pt}
    },
    outbox/.style={
        basebox, draw=cOut, fill=cOut!18, text=cOut!80!black,
        font=\LARGE\bfseries,
        minimum width=36mm, minimum height=16mm
    },
    flow/.style={
        -{Stealth[length=3.5mm, width=2.5mm, round]},
        line width=2.5pt, draw=cGray!85,
        rounded corners=6pt,
        line cap=round, line join=round
    },
    flowgood/.style={flow, draw=cOut},
    flowwarn/.style={flow, draw=cWarn},
    flowdist/.style={flow, draw=cDist},
    loopback/.style={
        -{Stealth[length=4mm, width=3mm, round]},
        line width=3.5pt, draw=cGen,
        line cap=round, line join=round
    },
    labeltext/.style={
        font=\LARGE\sffamily\bfseries,
        text=black!70,
        fill=white,
        rounded corners=4pt,
        draw=black!15, line width=0.8pt,
        inner xsep=6pt, inner ysep=3pt,
        align=center
    },
    badge/.style={
        circle,
        draw=cRes!95!black, fill=cRes!20,
        line width=1.4pt,
        inner sep=1.7pt,
        font=\LARGE\bfseries,
        text=black!75
    }
]

\node[solver] (S1) at (0,0) {$S_1$};
\node[solver] (S2) at (0,-17mm) {$S_2$};
\node[solver] (S3) at (0,-34mm) {$S_3$};
\node[solver] (S4) at (0,-51mm) {$S_4$};

\node[grader] (G1) at (55mm,0) {$G_1$};
\node[grader] (G2) at (55mm,-17mm) {$G_2$};
\node[grader] (G3) at (55mm,-34mm) {$G_3$};
\node[grader] (G4) at (55mm,-51mm) {$G_4$};

\coordinate (MG-1-1) at (S1);
\coordinate (MG-2-1) at (S2);
\coordinate (MG-3-1) at (S3);
\coordinate (MG-4-1) at (S4);
\coordinate (MG-1-2) at (G1);
\coordinate (MG-2-2) at (G2);
\coordinate (MG-3-2) at (G3);
\coordinate (MG-4-2) at (G4);

\begin{scope}[on background layer]
  \node[stagebox=cGen,   fit=(S1)(S4)] (pbox) {};
  \node[stagebox=cGrade, fit=(G1)(G4)] (gbox) {};
\end{scope}

\node[stagelabel=cGen,  anchor=south] (lblSolvers) at ([yshift=2pt]pbox.north) {Solvers};
\node[stagelabel=cGrade,anchor=south] (lblGraders) at ([yshift=2pt]gbox.north) {Grader copies};

\begin{scope}[on background layer]
  \node[rectangle, rounded corners=12pt,
        draw=cRes!95!black, dashed, dash pattern=on 7pt off 4pt,
        line width=2.6pt,
        fit=(pbox)(gbox)(lblSolvers)(lblGraders),
        inner sep=8pt] (outerbox) {}; 
\end{scope}

\node[basebox, draw=black!45, fill=white, left=14mm of pbox.west,
      minimum width=20mm, minimum height=14mm, font=\LARGE\bfseries] (problem) {Problem};
\draw[flow] (problem.east) -- (pbox.west);

\draw[flow] ([yshift=1.5mm]S1.east) -- ([yshift=1.5mm]G1.west);
\draw[flow, dashed, draw=cGray!60] ([yshift=-1.5mm]G1.west) -- ([yshift=-1.5mm]S1.east);
\draw[flow] ([yshift=1.5mm]S2.east) -- ([yshift=1.5mm]G2.west);
\draw[flow, dashed, draw=cGray!60] ([yshift=-1.5mm]G2.west) -- ([yshift=-1.5mm]S2.east);
\draw[flow] ([yshift=1.5mm]S3.east) -- ([yshift=1.5mm]G3.west);
\draw[flow, dashed, draw=cGray!60] ([yshift=-1.5mm]G3.west) -- ([yshift=-1.5mm]S3.east);
\draw[flow] ([yshift=1.5mm]S4.east) -- ([yshift=1.5mm]G4.west);
\draw[flow, dashed, draw=cGray!60] ([yshift=-1.5mm]G4.west) -- ([yshift=-1.5mm]S4.east);

\node[badge, anchor=north west] at ([xshift=7pt,yshift=-7pt]outerbox.north west) {A};

\node[decidebox, right=16mm of gbox.east] (decide) {Stalled?}; 
\node[outbox,    right=14mm of decide.east] (output) {Final\\Solution}; 

\draw[flow] (G1.east) -- ++(6mm,0) |- (decide.west);
\draw[flow] (G2.east) -- ++(6mm,0) |- (decide.west);
\draw[flow] (G3.east) -- ++(6mm,0) |- (decide.west);
\draw[flow] (G4.east) -- ++(6mm,0) |- (decide.west);

\draw[flowgood] (decide.east) -- node[labeltext, above, yshift=5pt] {no} (output.west);

\node[proc,  below=38mm of decide.south] (extract) {Conjecture\\Extractor};

\node[basebox, draw=cDist, fill=cDist!11, dashed,
      below=8mm of extract, minimum width=65mm, minimum height=12mm] (pool)
      {$C_1, \neg C_1, C_2, \neg C_2, \ldots$};

\node[resbox, below=8mm of pool] (resolver) {Conjecture Resolution\\[-2pt]{\large (solver + grader)}};

\begin{scope}[on background layer]
  \node[stagebox=cDist, fit=(extract)(pool)(resolver), inner sep=9pt] (cbox) {}; 
\end{scope}
\node[stagelabel=cDist, anchor=south] at ([yshift=2pt]cbox.north) {Conjecture Extraction};

\draw[flowdist]
  (decide.south)
  -- ++(0,-6mm)
  -| node[labeltext, right, pos=0.25, xshift=6pt] {yes}
  (extract.north);

\draw[flowdist] (extract.south) -- (pool.north);
\draw[flowdist] (pool.south) -- (resolver.north);

\node[badge, anchor=south west] at ([xshift=7pt,yshift=7pt]resolver.south west) {A};

\coordinate (loopEnd)   at ([yshift=-6mm]pbox.west);
\coordinate (loopStart) at (cbox.west); 

\draw[loopback]
  (loopStart)
  .. controls ($(loopStart)+(-40mm, 0mm)$) and ($(loopEnd)+(-40mm,-18mm)$)
  .. (loopEnd);

\node[labeltext, anchor=east] at ($(loopStart)+(-18mm, 0mm)$)
  {+ proof for conjecture\\or its negation};

\end{tikzpicture}%
}
      }
  \caption{A high level overview of our pipeline. A problem is processed by parallel solver--grader pairs (module \textbf{A}). When progress stalls, the \emph{Conjecture Extractor} identifies candidate lemmas $C_i$ and their negations from current proof attempts. Each conjecture is resolved by an independent solver--grader module (also \textbf{A}), with proofs (for either the conjecture or its negation) fed back to enable further progress toward the final solution.}
  \label{fig:v19-pipeline}
\end{figure}

\label{ap:alg_details}
\cref{alg:main_pipeline} is our main pipeline algorithm. Sub-functions are defined in other algorithms.

\begin{algorithm}[ht]
  \caption{Memory-Augmented Dialectic Reasoning (Main Pipeline)}
  \label{alg:main_pipeline}
  \begin{algorithmic}[1]
  \STATE \textbf{Input:} Problem $P$, Initial Iterations $L_0$, Conjecture Iterations $L$, Solvers $K$, Threshold $\tau=7$, Enhancement Threshold $\tau_{e} = 6$
  \STATE \textbf{Initialize:} Solution Memory $\mathcal{M}_{sol} \leftarrow \emptyset$, Lemma Memory
  $\mathcal{M}_{lemma} \leftarrow \emptyset$, Failure Context $\mathcal{C}_{fail} \leftarrow \emptyset$

  \STATE \textsc{// Phase 1: Baseline Exploration with Feedback Refinement}
  \STATE $\mathcal{S}_{base} \leftarrow \emptyset$ 
  \FOR{$i = 1$ \textbf{to} $L_0$}
      \FOR{$k = 1$ \textbf{to} $K - 1$}  
      \STATE $Ctx_k \leftarrow \textsc{SelectKthTop}(\mathcal{S}_{base},k)$
      \ENDFOR
      \STATE $\mathcal{S}_{base} \leftarrow \mathcal{S}_{base}\cup \textsc{DialecticSolve}(P, \emptyset) \,\cup$
      \item[]\;\;\;\;\;\;\;\;\; $ \bigcup_{k \le K - 1} \textsc{DialecticSolve}(P, Ctx_k)  $
      \IF{$\textsc{VerifiedSuccess}(\mathcal{S}_{base}, N=3)$}
          \STATE \textbf{return} $\textsc{Best}(\mathcal{S}_{base})$
      \ENDIF
  \ENDFOR
  \STATE $\mathcal{M}_{sol} \leftarrow \textsc{Update}(\mathcal{M}_{sol}, \mathcal{S}_{base})$

  \FOR{$l = 1$ \textbf{to} $L$}
      \STATE \textsc{// Phase 2: Learning from Failure}
      \STATE $S_{seeds} \leftarrow \textsc{SelectTop}(\mathcal{M}_{sol}, N=2)$
      \STATE $\mathcal{H} \leftarrow \textsc{ExtractHypotheses}(S_{seeds}, \mathcal{M}_{lemma},
  \mathcal{C}_{fail})$
      \STATE $\mathcal{L}_{new}, \mathcal{F}_{new} \leftarrow \textsc{VerifyHypotheses}(\mathcal{H}, \tau)$
      \STATE $\mathcal{M}_{lemma} \leftarrow \mathcal{M}_{lemma} \cup \mathcal{L}_{new}$
  \STATE $\mathcal{C}_{fail} \leftarrow \mathcal{C}_{fail} \cup \mathcal{F}_{new}$

      \STATE \textsc{// Phase 3: Solving with Enhanced Context}
      \STATE $Ctx \leftarrow \mathcal{M}_{lemma} \cup \textsc{RetrieveBest}(\mathcal{M}_{sol})$
      \STATE $\mathcal{S}_{guided} \leftarrow \textsc{DialecticSolve}(P, Ctx, K-1)$
      \STATE $\mathcal{S}_{fresh} \leftarrow \textsc{DialecticSolve}(P, \emptyset, 1)$ \COMMENT{Diversity
  check}

      \STATE $\mathcal{S}_{final} \leftarrow \mathcal{S}_{guided} \cup \mathcal{S}_{fresh}$
      \STATE $\mathcal{M}_{sol} \leftarrow \textsc{Update}(\mathcal{M}_{sol}, \mathcal{S}_{final})$

      \IF{$\textsc{VerifiedSuccess}(\mathcal{S}_{final}, N=3)$}
          \STATE \textbf{return} $\textsc{Best}(\mathcal{S}_{final})$
      \ENDIF
  \ENDFOR

  \STATE \textsc{// Phase 4: Post-Enhancement for Near-Perfect Solutions}
  \STATE $S^* \leftarrow \textsc{Best}(\mathcal{M}_{sol})$
  \STATE $\mathcal{G}_{check} \leftarrow \textsc{GradeIndependent}(S^*, N=3)$
  \IF{\textbf{not} $\textsc{AllPerfect}(\mathcal{G}_{check})$} 
      \STATE $\mathcal{H}_{gaps} \leftarrow \textsc{ExtractGaps}(S^*, \mathcal{G}_{check})$
      \STATE $\mathcal{L}_{fix}, \mathcal{F}_{fix} \leftarrow \textsc{VerifyHypotheses}(\mathcal{H}_{gaps}, \tau_e)$

      \STATE $S_{better} \leftarrow \textsc{DialecticSolve}(P, \mathcal{L}_{fix} \cup
  \mathcal{F}_{fix}, K)$
      \STATE $\mathcal{G}_{better} \leftarrow \textsc{GradeIndependent}(S_{better},N=3)$
	  \STATE \textbf{return} $\text{argmax}\{\textsc{MEAN}(\mathcal{G}_{check}), \textsc{MEAN}(\mathcal{G}_{better})\}$
  \ENDIF
  \STATE \textbf{return} $S^*$
  \end{algorithmic}
  \end{algorithm}
  
There are further details on sub-functions in Algorithm \ref{alg:dialectic_solve}. Note that Algorithm \ref{alg:dialectic_solve} is orchestrator-level pseudocode: the parallel threads and the separate LAZYPHRASING model call (Gemini 2.5 Flash) represent distinct API calls managed by the Python orchestrator, not steps within a single prompt execution.

\begin{algorithm}[ht]
\caption{Function: DialecticSolve}
\label{alg:dialectic_solve}
\begin{algorithmic}[1]
\STATE \textbf{Input:} Problem $P$, Context $Ctx$, Count $N=1$
\STATE $\mathcal{S}_{out} \leftarrow \emptyset$

\FOR{$i = 1$ \textbf{to} $N$ \textbf{in parallel}}
    \STATE \textsc{// Step 1: Draft \& Check}
    \STATE $S_i \leftarrow \text{LLM}_{\text{solve}}(P, Ctx)$
    \IF{$\textsc{LazyPhrasing}(S_i)$}
        \STATE $S_i \leftarrow \text{LLM}_{\text{solve}}(P, Ctx,$
        \STATE \quad \quad $\text{feedback=``Derive explicitly"})$
    \ENDIF
    
    \STATE \textsc{// Step 2: Inquisitorial Grading}
    \STATE $F_i \leftarrow \text{LLM}_{\text{grader}}(S_i)$
    \STATE $S'_i \leftarrow \text{LLM}_{\text{refine}}(S_i, F_i)$
    \STATE \textsc{// Step 3: Re-Grade Refined Solution}
    \STATE $F'_i \leftarrow \text{LLM}_{\text{grader}}(S'_i)$
    \STATE $\mathcal{S}_{out} \leftarrow \mathcal{S}_{out} \cup \{(S'_i, F'_i)\}$ 
\ENDFOR
\STATE \textbf{return} $\mathcal{S}_{out}$
\end{algorithmic}
\end{algorithm}

\begin{algorithm}[ht]
\caption{Function: VerifyHypotheses}
\label{alg:verify_hypotheses}
\begin{algorithmic}[1]
\STATE \textbf{Input:} Candidates $\mathcal{H}$, Threshold $\tau$
\STATE $\mathcal{L}_{proven} \leftarrow \emptyset, \quad \mathcal{F}_{failed} \leftarrow \emptyset$
\FORALL{$(C, \neg C) \in \mathcal{H}$}
    \STATE $g_{pos} \leftarrow \textsc{Eval}(\text{LLM}_{\text{solve}}(C))$
    \STATE $g_{neg} \leftarrow \textsc{Eval}(\text{LLM}_{\text{solve}}(\neg C))$
    
    \IF{$g_{pos} \ge \tau \land g_{neg} < \tau$}
        \STATE $\mathcal{L}_{proven} \leftarrow \mathcal{L}_{proven} \cup \{C\}$
    \ELSIF{$g_{neg} \ge \tau \land g_{pos} < \tau$}
        \STATE $\mathcal{L}_{proven} \leftarrow \mathcal{L}_{proven} \cup \{\neg C\}$
    \ELSE
        \STATE $\mathcal{F}_{failed} \leftarrow \mathcal{F}_{failed} \cup \{(C, \text{Ambiguous})\}$
    \ENDIF
\ENDFOR
\STATE \textbf{return} $\mathcal{L}_{proven}, \mathcal{F}_{failed}$
\end{algorithmic}
\end{algorithm}

\begin{algorithm}[ht]
\caption{Parallel Runs}
\label{alg:parallel_runs}
\begin{algorithmic}[1]
 \STATE \textbf{Input:} Problem $P$, Number Parallel Generations $N=2$
 \FOR{$i = 1$ \textbf{to} $N$ \textbf{in parallel}}
    \STATE $S_i \gets \textsc{MemoryAugmentedDialectic}(P)$
 \ENDFOR
 \STATE \textbf{return} \textsc{PickBest}($S_1, \dots, S_N$)
 
\end{algorithmic}
\end{algorithm}

\section{Ablations}
In this section we track various ablations over benchmarks, graders, and hyperparameters among others. 

\subsection{Other Benchmarks}
\label{ap:other_benchmarks}
\begin{figure}[h]
    \centering
    \includegraphics[width=0.75\linewidth]{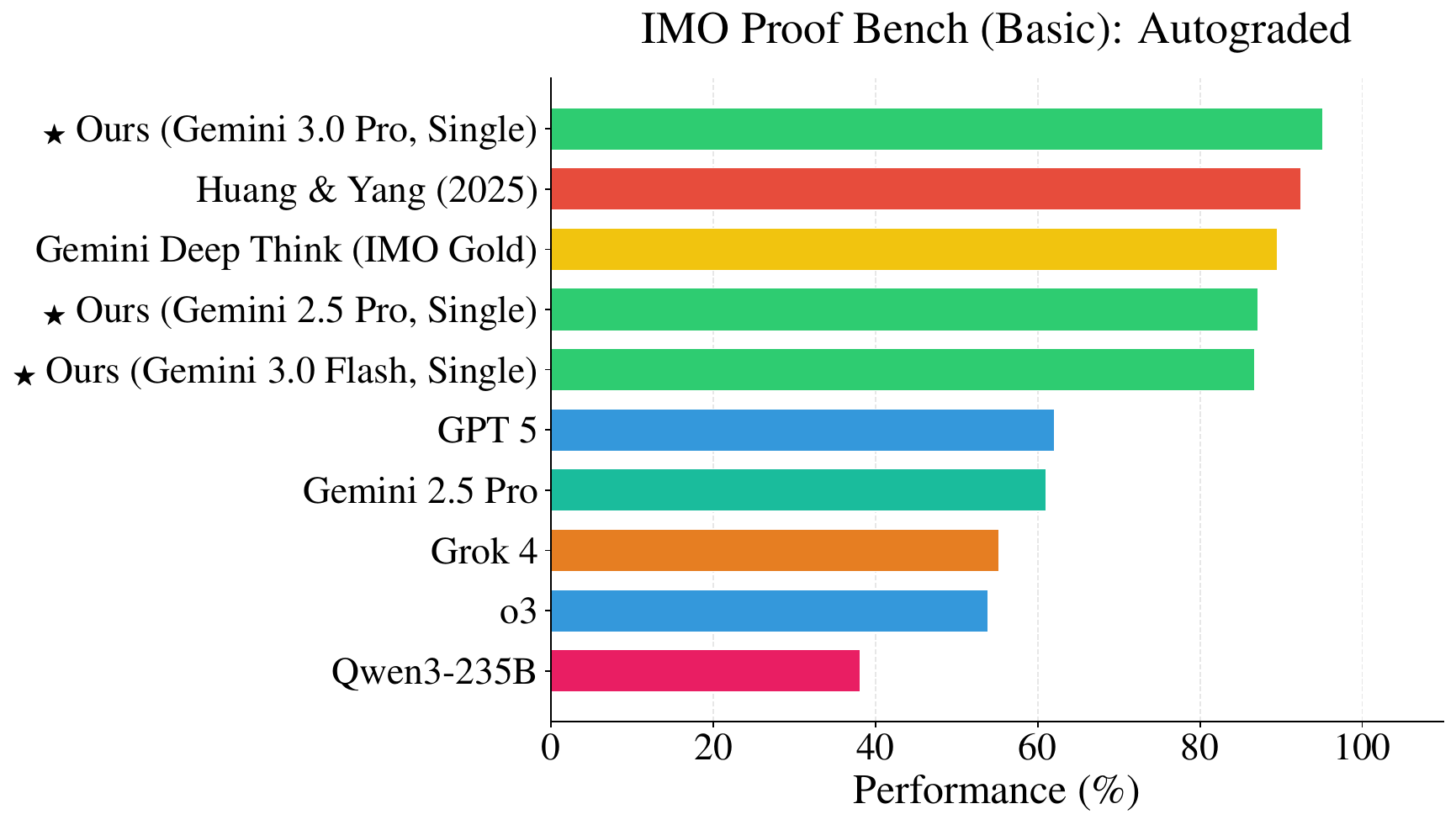}
    \caption{Results for IMO Proof Bench (Basic) according to the IMO-Bench autograder. ``Single" here denotes that the output of a \emph{single} run of the pipeline was submitted (as opposed to our standard algorithm for PB-Adv which uses a further model call to choose from two parallel pipeline runs - cf. Section \ref{subsec:pipeline_arch} ``Post Enhancement and Aggregation"). All comparison results taken from \citet{luong2025towards}.}
    \label{fig:pb_basic}
\end{figure}

While PB-Adv is the most challenging area to measure the performance of models, PB-Basic  also presents useful signal for practitioners seeking to evaluate capabilities on slightly less challenging problems. We demonstrate the success of our model on PB-Basic  in Figure \ref{fig:pb_basic}. We see that our pipeline -- run only once using several architectural variants, achieves competitive scores with the top pipelines. Since it is allowed to converge early, we incur minimal inference costs, for example $4.54$ USD/question for 3.0 Pro, and $1.23$ USD/question for 3.0 flash.

In addition, we report performance on two more benchmarks: ProofBench - a similarly named but independent benchmark of roughly $400$ questions from a variety of sources including the Putnam exam, and IMOAnswerBench - a final answer only benchmark comprising a larger number of competition math questions.

\begin{table}[h]
\centering
\caption{Results on ProofBench and IMO-AnswerBench. Our pipeline runs on Gemini 3.0 Flash; baselines from QED-Nano.}
\label{tab:more_benchmarks}
\begin{tabular}{lcc}
\toprule
\textbf{Model} & \textbf{ProofBench} & \textbf{IMO-AnswerBench} \\
\midrule
Qwen3-30B-A3B-Thinking-2507  & 26.1 & 67.0 \\
Qwen3-235B-A22B-Thinking-2507 & 33.7 & 70.5 \\
GPT-OSS-120B                  & 47.5 & 70.5 \\
DeepSeek-Math-V2              & 60.6 & 75.8 \\
Gemini 3 Pro                  & 66.7 & 83.2 \\
\textbf{Ours (Gemini 3.0 Flash)} & \textbf{80.3} & \textbf{91.3} \\
\bottomrule
\end{tabular}
\end{table}

Our pipeline outperforms all baselines on both benchmarks despite using our weakest backbone. Note that broader evaluations remain cost-prohibitive at academic budgets: running our pipeline on the $400$ IMO-AnswerBench problems with Flash alone costs roughly \$$1{,}000$.

\subsection{Backbone Ablation} \label{ap:backbone}
Here we show gains on PB-Adv using different backbone models for each of our pipeline roles. 

\begin{table}[h]
\centering
\caption{Generalization across backbones on PB-Advanced. ``Base Avg'' is the autograded average score (\%) for a single call to the model; ``Ours Avg'' is the score of our pipeline using that model as the backbone. ``Graded (Ours)'' is the number of problems for which grading had completed at the time of writing.}
\label{tab:cross_backbone}
\begin{tabular}{lcc}
\toprule
\textbf{Backbone} & \textbf{Base Avg (\%)} & \textbf{Ours Avg (\%)}\\
\midrule
Claude 4.6 Opus    & 23.8 & 85.7 \\
DeepSeek-V3.2      & 15.2 & 66.7 \\
\bottomrule
\end{tabular}
\end{table}
\subsection{Statistical Significance}\label{subsec:significance}

Significance testing in this setting is challenging because two sources of randomness are at play: prover sampling and LLM-grader sampling. Hiring expert human graders at scale to mitigate the latter is financially infeasible at academic budgets. To bound prover variance we ran $10$ independent trials on Gemini 3.0 Flash---our cheapest backbone---on PB-Advanced; running the same number of trials on Gemini 2.5 Pro would have cost over \$$12{,}000$. These trials yield a Student-$t$ $95\%$ confidence interval on the mean autograded score of $[49.1\%, 56.1\%]$ on PB-Advanced. Even the lower bound exceeds the autograded score reported by \citet{huang2025winning}, despite our pipeline using a weaker backbone and a budget orders of magnitude smaller.

\subsection{\citet{huang2025winning} Grader Analysis}\label{ap:grader_analysis}
Due to resource constraints, we were unable to run the expensive \citet{huang2025winning} pipeline more than a few times on PB-Adv. We focus on a single run, using the default settings and graded using ProofAutoGrader, which does not reveal all of the false positives that \citet{luong2025towards}  observe. The results of grading this run is summarized in Table~\ref{tab:hy_grader_checked_by_pag}.
\begin{table}[h]
\centering
\caption{The counts of how many questions received each possible autograder result on \citet{huang2025winning} pipeline solutions. None of the proposed solutions earned a 1 through 6. Note that the pipeline does not finish (DNF) if it does not grade a solution correct $5$ times in a row.}
\label{tab:hy_grader_checked_by_pag}
\begin{tabular}{cc}
\toprule
\textbf{ProofAutoGrader Score} & \textbf{Count}\\
\midrule
0/7 &  1  \\
7/7 & 14 \\
DNF & 15 \\
\bottomrule
\end{tabular}
\end{table}

The only false positive we observed in this run is \textbf{PB-Adv-011}. For completeness, the full solution text is in Appendix~\ref{ap:pb-011-res}.

We study this case further in Table~\ref{tab:grader_comparison_appendix} by running 20 independent verifications with the \citet{huang2025winning} verifier on the following solutions:
\begin{itemize}
\item  The solution output by the \citet{huang2025winning} pipeline (that was verified 5 times in a row).
\item The solution output by the pipeline, without the following lines:
“I have not found a complete solution. However, I have rigorously proven two significant results: 1. The only possible injective solution is the function $f(x) = 1/x$. 2. Any solution $f$ for which $f(1) \neq 1$ must be non-injective. Combining these results, the problem is reduced to proving that no non-injective solutions exist. I have been unable to rigorously prove this final step." 
\end{itemize}
\begin{table}[h]
\centering
\caption{Huang \& Yang Verifier run 20 times independently on the solution the pipeline emitted on \textbf{PB-Adv-011}. We see that truncation gets rid of the false positives.}
\label{tab:grader_comparison_appendix}
\begin{tabular}{lc}
\toprule
\textbf{Response} & \textbf{Accepted by Huang \& Yang Verifier}  \\
\midrule
\textbf{Full} & $6/20$\\
\textbf{Truncated} & $0/20$\\
\bottomrule
\end{tabular}
\end{table}

Our conclusion is thus when given an admission that the solution is incomplete, this grader thinks only to check if the lemmas are true, not the overall statement. This is indicative of bad prompt engineering or poor instruction-following from Gemini. Either way, this is a \textbf{Cognitive Well} of the model--the solver was able to ``reward-hack'' its way into a solution that was amenable to the grader.

\subsection{ Conjecture Removal: Further Analysis}
\label{ap:conj_removal_details}

\begin{figure*}[h]

    \includegraphics[width=\textwidth]{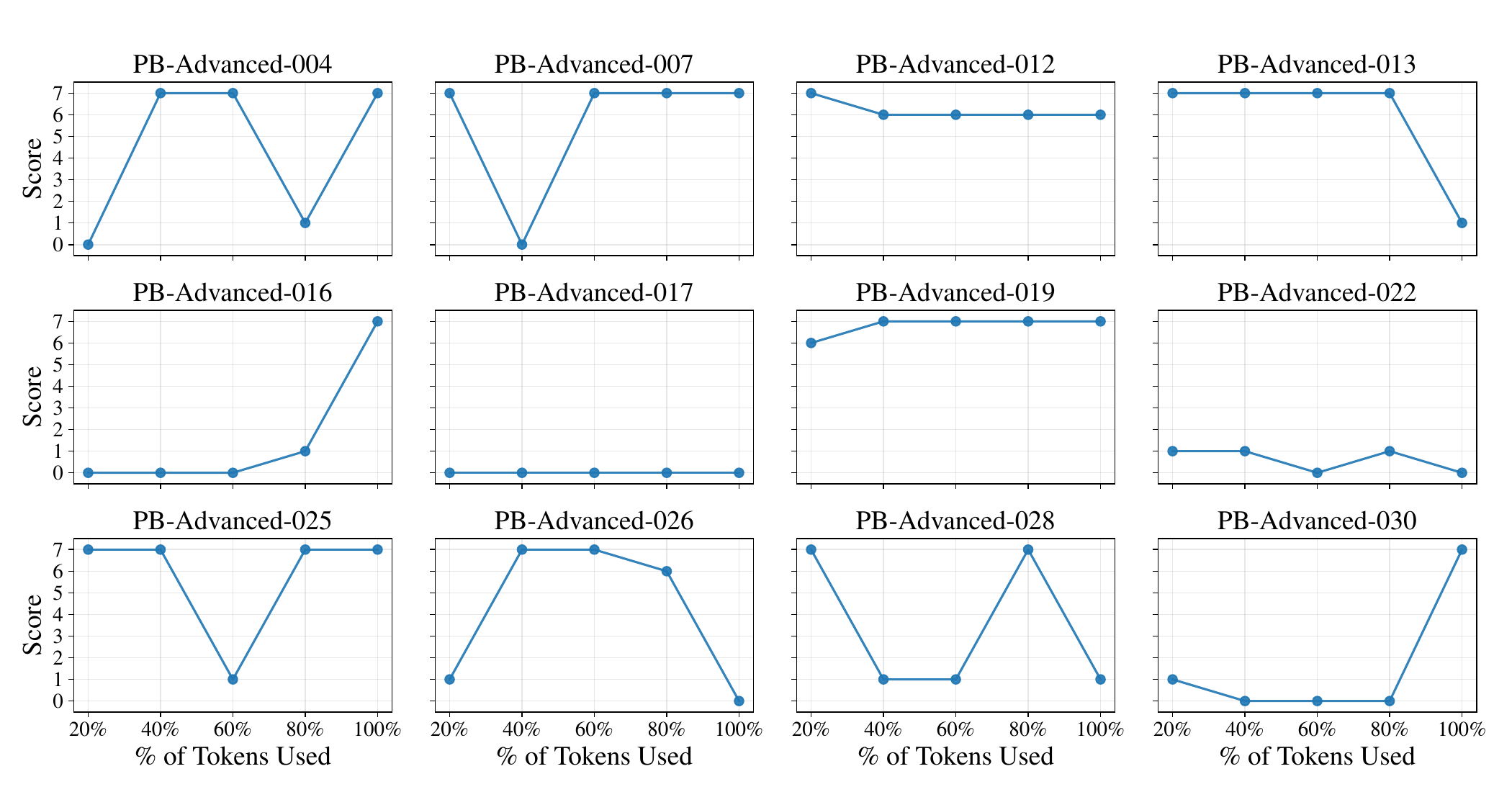}
    \caption{Effect of removing conjectures from successful runs of the solver pipeline, as described in \cref{subsec:conj_removal}.  The $x$-axis denotes progression in time and the $y$-axis corresponds to the grade provided by ProofAutoGrader to the pipeline's best solution. After removal of conjectures, the solution process becomes erratic or confused. Compare to Figure~\ref{fig:phase1_focused_traj}}
    \label{fig:phase1_trajectories}
\end{figure*}

\cref{fig:phase1_trajectories}
shows a broader selection of trajectories (all of which originally scored $7/7$); $6$ of them ended with a score below $7$ without conjectures, and all $6$ ended at a score of exactly $0$. We can see in this case that the Phase 1 refinement process is quite noisy --solutions bounce between scores throughout the run, with both local increases and decreases. This bolsters intuition that our conjecture proving rounds and lemma pool can stabilize the iterative proof process and may prevent negative progress that would otherwise lead the pipeline to emit worse proofs. 

\subsection{Momus Solver Evaluation: Single Call}
\label{ap:solver_ablation}

In Section \ref{results:solver}, we compared two solver configurations to isolate the effect of prompt engineering:
\begin{itemize}
    \item \textbf{Momus Solver}: A  single call to our dialectic-based solver prompt. This is captured in Appendix~\ref{prompt:solver-resurrected}.\footnote{This is a slight variant of the  prompt in Appendix~\ref{prompt:solver}. }
    \item \textbf{Huang Yang Solver}: A single call using the Solver prompt in \citet{huang2025winning}.
\end{itemize}

For each of the 30 Proofbench-Basic problems, we generated 4 independent solution attempts using each configuration with Gemini 2.5 Pro. All solutions were scored using the \textbf{ProofAutoGrader} \citep{luong2025towards}. Table~\ref{tab:solver_comparison_appendix} details the head-to-head performance.

\begin{table}[h]
\centering
\caption{Comparison of one call to Momus Solver vs. Huang Yang Solver using Average Scores on Proofbench-Basic (30 problems, 4 attempts each). ``Score @ 1'' is the score on the first attempt (out of 7); ``Max Score @ 4'' is the highest score across all four attempts.}
\label{tab:solver_comparison_appendix}
\begin{tabular}{lcc}
\toprule
\textbf{Metric} & \textbf{HY Average} & \textbf{Momus Average}\\
\midrule
Score @ 1 & 4.90 & 5.33 \\
Max Score @ 4  & 6.50 & 6.67 \\
Average-of-4 & 4.92 & 5.10 \\
\bottomrule
\end{tabular}
\end{table}

\subsection{Width ($K$) and Depth ($L$) Sweep}\label{ap:kl_sweep}

To complement the ablations in the main text, we report a sweep over the two main hyperparameters of our pipeline: the parallel solver width $K$ and the number of conjecture-iteration outer rounds $L$. Extended ablations of this form are costly (more than \$$100$/run), so we conduct the sweep on Gemini 3.0 Flash---our cheapest backbone---on PB-Advanced. Tables~\ref{tab:l_sweep} and \ref{tab:k_sweep} report autograded average scores.

\begin{table}[h]
\centering
\caption{Depth ($L$) sweep at fixed $K=4$ on Gemini 3.0 Flash, PB-Advanced.}
\label{tab:l_sweep}
\begin{tabular}{lc}
\toprule
\textbf{Config} & \textbf{Avg (\%)} \\
\midrule
$L=1$, $K=4$ & 52.4 \\
$L=3$, $K=4$ & 58.1 \\
$L=6$, $K=4$ & 47.6 \\
\bottomrule
\end{tabular}
\end{table}

\begin{table}[h]
\centering
\caption{Width ($K$) sweep at fixed $L=3$ on Gemini 3.0 Flash, PB-Advanced.}
\label{tab:k_sweep}
\begin{tabular}{lc}
\toprule
\textbf{Config} & \textbf{Avg (\%)} \\
\midrule
$L=3$, $K=2$ & 59.6 \\
$L=3$, $K=4$ & 58.1 \\
$L=3$, $K=8$ & 60.5 \\
\bottomrule
\end{tabular}
\end{table}

We do not observe a strong monotonic trend in either sweep, plausibly because run-to-run noise on a $30$-problem benchmark is non-negligible at this scale. Importantly, every configuration in the sweep remains highly competitive on PB-Advanced, suggesting that the pipeline is fairly robust to these hyperparameter choices and does not require fine grid-tuning to achieve strong performance.

\subsection{Context Contamination Ablation}
\label{ap:context_contamination}

A central claim of our work is that the ``Cognitive Well'' is partially driven by \emph{context contamination}: the solver's and grader's judgment on a sub-claim is degraded by the surrounding flawed proof, which is precisely what motivates contextual detachment in our pipeline. To test this directly, we ran a controlled ablation on the conjecture-proving step. We parsed every extracted conjecture from a full Gemini~3 Flash Preview Momus run on ProofBench Advanced (parallel depth $3$, Phase~I width $4$), comprising $N{=}73$ verification instances across all $30$ advanced problems, along with each instance's isolated verification outcome as logged by the deployed pipeline. We then re-ran the same prover/grader stack on those same statements under two conditions: (a) in isolation (the deployed setting), and (b) with the original problem statement and two Phase~I solver transcripts from that problem's parent run (the last two Step~1 / feedback-injection drafts recovered from the run's API log) supplied as additional context, together with instructions that the lemma was extracted while attempting the original problem.

The re-proving experiment used the same internal dialectic solver prompt, Momus grader prompt, grading extraction rule, and final-blueprint extraction used by the pipeline; success meant that either the conjecture or its negation received a verifier score at least $5/7$, matching the pipeline's conjecture-verification threshold. Solver and grader both used temperature $1.0$, max output length $65{,}536$, and high thinking level. Aggregating over all $73$ instances, isolated verification succeeded $30/73$ times ($41.1\%$), whereas with parent-run context it succeeded only $25/73$ times ($34.2\%$), i.e.\ an absolute drop of $6.9$ percentage points ($\approx\!17\%$ relative to the isolated rate). More sharply, among the $n{=}30$ instances that verified in isolation, $20$ failed to verify after appending parent context---a \textbf{20/30 $\approx$ 67\%} \emph{failure rate on previously successful lemmas}. These findings support contextual detachment: the model's ability to settle a sub-claim is materially harmed by proximity to the original solve attempt. We did \emph{not} include a length-matched neutral distractor context; accordingly, we cannot fully separate logical contamination from generic effects of processing substantially longer prompts (an informative control, but one we leave for future work).

\section{Compute details}
\subsection{Mechanical Details}

Unless otherwise stated, our pipeline exclusively employs \textbf{Gemini} \citep{gemini3announce2025, comanici2025gemini} models as the backbone for Solver, Grader, and Extractor modules, utilizing a temperature of $T=0.6$ for generation and $T=0.1$ for verification for $2.5$ Pro and $T=1.0$ for $3.0$ variants and roles. We utilize \textbf{Gemini 2.5 Flash} solely for the Answer-Processor (\texttt{LazyPhrasing}) to reduce latency.

The ``Conjecture Extraction" phase operates on a fixed budget: we extract the top $k=3$ most ``load-bearing" sentences from a candidate proof. The Bisection Test (proving $C$ and $\neg C$) is run in parallel threads. To prevent infinite loops, we cap the Global Memory Refinement at 3 outer iterations.

\subsection{Cost of Our Method}
\label{ap:compute_ours}
In Table \ref{tab:cost_breakdown} present more detailed analysis of the compute used to achieve our performance numbers on IMO-ProofBench Advanced.

The average researcher is  unable to afford large expensive orchestrations. Our efficient method presents an especially attractive option for experiments  on proof-based math problems. To demonstrate this, we first document our total costs in Table \ref{tab:cost_breakdown}. Compared to straightforward calculations and  implementations of other methods (see Appendix \ref{ap:compute_competitors}), our solver system saves close to $90\%$ of the total cost compared to the next closest \citet{huang2025winning} pipeline while exceeding its performance on difficult PB-Adv questions. Furthermore, we demonstrate the success of our pipeline with newer, more performant models, as well as more affordable offerings and show in Figure \ref{fig:pareto} that our method dominates the Pareto frontier of advanced questions --demonstrating capable performance for as little as $3$ USD/question!

\begin{table}[ht]
    \centering
    \caption{Cost Breakdown (USD) by model for all PB-Adv questions.}
    \label{tab:cost_breakdown}
    \begin{tabular}{lr}
        \toprule
        \textbf{Run} & \textbf{Total Cost} \\
        \midrule
        Gemini 2.5 Pro Run 1 & \$607.60 \\
        Gemini 2.5 Pro Run 2 &\$618.88 \\
        \textbf{Gemini 2.5 Pro Combined} & \textbf{\$1,226.48} \\
        \midrule
        Gemini 3.0 Pro Run 1 &\$428.16 \\
        Gemini 3.0 Pro Run 2 & \$505.48 \\
        \textbf{Gemini 3.0 Pro Combined} & \textbf{\$933.64} \\
        \midrule
        Gemini 3.0 Flash Run 1 & \$93.60 \\
        Gemini 3.0 Flash Run 2 & \$89.86 \\
        \textbf{Gemini 3.0 Flash Combined} & \textbf{\$183.46} \\
        \bottomrule
    \end{tabular}
\end{table}

\subsection{Comparisons}
\label{ap:compute_competitors}
Explicit budgets and token counts for results on PB-Adv are not generally released. However, based on pipeline descriptions, iteration counts, and parallelism, we derive here rough estimates for the tokens and cost for other pipelines.

For \citet{huang2025winning} we calculate maximum budget based explicitly on the pipeline described in \citet{luong2025towards} which describes: 
\begin{quote}
thinking budget (32K tokens) per model call ... [with] repeated iterations (at most 30) of “selfverification” and “bug-fixing” on it ... This entire pipeline will be run in parallel multiple times (100) as well, until there is at least one solution returned from any run. Theoretically the model could fail to find any solution after all
parallel runs, which occurred for two IMO-Proof Bench (Advanced) problems.
\end{quote}

Which amounts to a (roughly estimated) maximum budget of $32,000 \text{ tokens/call} \times 30 \text{ rounds/pipeline run} \times 2 \text{ calls/round} \times 100 \text{ pipeline runs/question} \times 10 \text{ USD/million tokens} = 1920 \text{USD/question}$.  Note this is a maximum budget, and in practice, we observe a more modest expense (extrapolated to $100$ iterations) of roughly $372$ USD/question.

For \citet{shao2025deepseekmath}, the pipeline employs a $128$K token limit for each LLM call. The pipeline starts with $64$ initial proofs $\times 64$ analyses per proof. Then, the top $64$ candidate proofs are selected to survive to the next round, and refined. This iterates for up to $16$ rounds. This leads to a conservative estimate of the compute limit as $16 \times 64 \times 64 > 64$K LLM calls. Together with the per-call token budgets, as well as the cost of the closest hosted model (DeepSeek 3.2 \citep{liu2025deepseek}), this amounts to an allowance of over $3000$ USD/question.

\begin{figure*}[h]
    \centering
    \includegraphics[width=\textwidth]{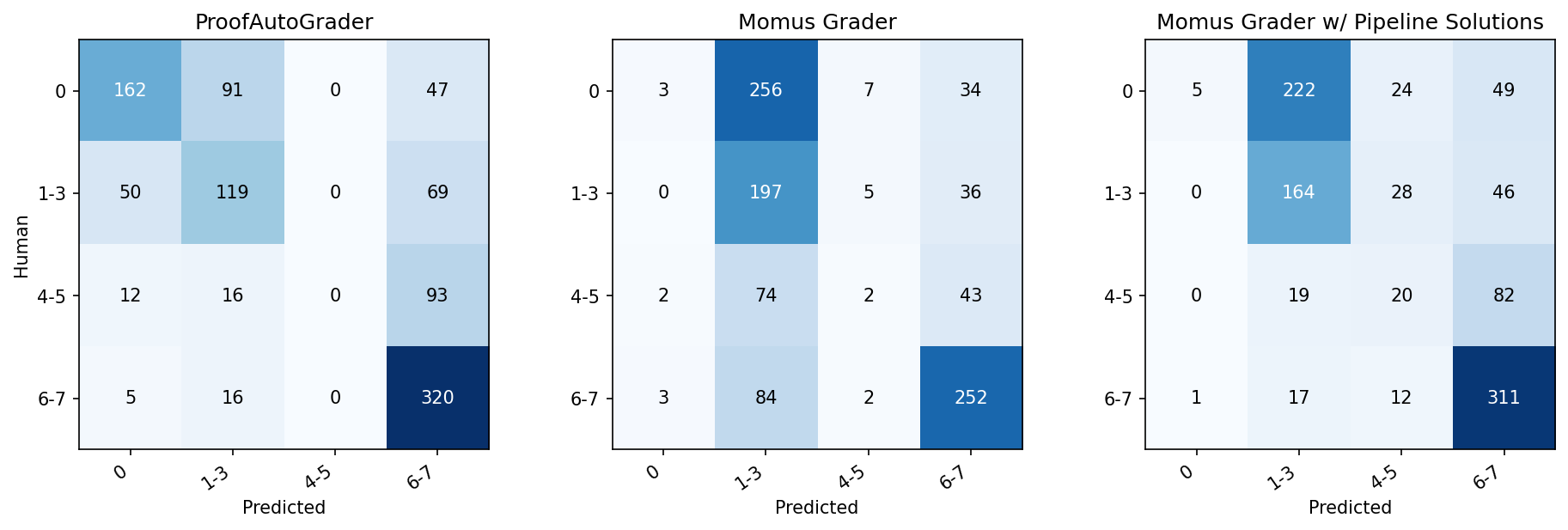}
    \caption{Confusion matrices of the two Momus graders against the autograder in \citet{luong2025towards}.}
    \label{fig:grader_confusion}
\end{figure*}

\section{Escaping the Cognitive Well: Qualitative analysis}
\label{ap:cognitive_well_analysis}
In this section we present three instances where the pipeline generated independent conjectures that were significant to produce a final correct solution. We present the conjectures and explain how they were useful to help the model generate the final solution. We observe these conjectures act as ``logical anchors" that reorient the search space toward a solvable path, effectively avoiding the Cognitive Well.

\subsection{PB-Advanced-019}

\begin{itemize}[nosep]
    \item \textbf{Problem Statement:} For a real number $r$, let $A(r)$ denote the fractional part of $2r$ in its decimal representation. For a real number $r$ and a positive integer $n$, define $B(n, r) = \sum_{k=1}^n A(kr)$. Find all positive real numbers $r$ such that $n(n + 1)r - B(n, r)$ is a multiple of $n$ for all positive integers $n$.
    
    \item \textbf{Conjecture:} The Gemini 2.5 pipeline independently identified the following lemma:
    \begin{equation*}
        \forall \alpha \in (0, 1), \exists n \in \mathbb{Z}_{\text{odd}} \text{ s.t. } n \nmid \sum_{k=1}^n \lfloor k\alpha \rfloor
    \end{equation*}
    
    \item \textbf{Utility:} This conjecture is the resulting statement needed to be proven after an algebraic simplification of the problem. This simplification for odd integers $n$ pivoted the solution towards the direction of bounding the sum of fractional parts, which ultimately led to the result that $r$ must be an integer.
\end{itemize}

\subsection{PB-Advanced-016}
    
\begin{itemize}[nosep]
    \item \textbf{Problem Statement:} Let $ABC$ be a non-isosceles triangle with incenter $I$. Let line $BI$ intersect $AC$ at $E$, and line $CI$ intersect $AB$ at $F$. Two points $U$ and $V$ are on segments $AB$ and $AC$ respectively, such that $AU = AE$ and $AV = AF$. Let the line passing through $I$ and perpendicular to $AI$ intersect line $BC$ at $L$. The circumcircle of $\triangle ILC$ intersects line $LU$ at $X$ (other than $L$), and the circumcircle of triangle $\triangle ILB$ intersects line $LV$ at $Y$ (other than $L$). Prove that if $P$ is the intersection of lines $YB$ and $XC$, then line $IP$ is parallel to line $XY$.

    \item \textbf{Conjecture:} The pipeline independently identified the lemma: $L, U, V \text{ are collinear}$ and correctly verified via Menelaus Theorem:
    \begin{align*}
         \iff\frac{AU}{UB} \cdot \frac{BL}{LC} \cdot \frac{CV}{VA} = 1
    \end{align*}

    \item \textbf{Utility:} This conjecture served as key logical step in the argument that significantly simplifies the proof. First, it simplifies by reducing lines $LU,LV, XY$ to a single line. Secondly, it allows the problem to be greatly transformed via inversion (since the inverse of points $X,Y$ become the intersections of the unique line through $L-U-V-X-Y$ and lines $IB$, $IC$. This enabled the model to relate the circumcircles of $\triangle ILC$ and $\triangle ILB$ via a common line $L-U-V$.

\end{itemize}

\subsection{PB-Advanced-003}
    
\begin{itemize}[nosep]
    \item \textbf{Problem Statement:} Let $ABC$ be an acute, non-isosceles triangle. Let $I$ be the incenter and let $\omega$ be the circumcircle of $ABC$. Let the intersections of lines $AI, BI,$ and $CI$ with sides $BC, CA,$ and $AB$ be $D, E,$ and $F$ respectively. 

Furthermore, let $\omega_A$ be the circle that lies inside $\angle BAC$, tangent to the lines $AB$ and $AC$, and internally tangent to the circumcircle $\omega$ at point $T_A$. Similarly, define points $T_B$ and $T_C$ as the points of tangency for circles $\omega_B$ and $\omega_C$ corresponding to vertices $B$ and $C$ respectively.

Prove that there exist two points $X$ and $Y$ such that the circumcircles of triangles $ADT_A$, $BET_B$, and $CFT_C$ all pass through $X$ and $Y$.

    \item \textbf{Conjecture:} The pipeline independently conjectured that $AT_A$, $BT_B$ and $CT_C$ are concurrent and correctly proved it through an application of Monge's Theorem.

    \item \textbf{Utility:} This conjecture is significant because it establishes that three radical axis of each pair of desired circles are concurrent. This effectively reduces the problem to proving that the 3 circles intersect at one point, instead of two. The pipeline then solves the problem using the concurrence as a key lemma.

\end{itemize}

\section{Limitations of Our Pipeline}
\label{ap:limitations}

Though our pipeline shows gains in efficiency and performance, several limitations warrant discussion. 

First: Consistent with prior work on math reasoning \citep{luong2025towards}, we employ an autograder for many of our ablations and comparisons.  While \citet{luong2025towards} reports high correlation on both basic ($r=0.98$) and advanced ($r=0.93$) between their autograder and human experts, they also document occasional unreliability  (e.g., $15\%$ difference on Deep Think (IMO Gold) pipeline!). We employed an expert human grader (provided a gold solution and rubric) for two of our most important pipeline runs, and found their grades consistent with the autograder. 

Second: lack of evaluation data of sufficient size and complexity. This is primarily due to 
contamination concerns and verification concerns.  While IMO-Bench is a significant improvement, its advanced evaluation subset comprises only $30$ problems.  To mitigate contamination effects we test on the same backbone models as previous papers (many  released \emph{prior} to  release of IMO-Bench).

Third: while our pipeline wins on cost and accuracy, it loses to current massively parallel pipelines on latency.

Finally, cost considerations prevented us from evaluating our pipeline using a wider set of frontier models and from trying extensive prompt tuning.

\section{System Prompts}
\label{sec:prompts}

\subsection{Notes on prompting framework (``Momus")}
\label{ap:momus}
 \noindent{\bf Dialectic-based prompting:} 
   Current pipelines (e.g. \citet{huang2025winning, ren2025deepseek}) prompt the model by carefully describing the task using more than a page of precise instructions.  Minor changes in such prompts  can lead to unpredictable, human-perceptible changes in outcome. Due to resource constraints, the number of pipelines runs we could run was fairly limited.
    This motivated us to employ a prompt-design framework optimized for predictable tuning rather than raw performance. While this approach yields minor (if any) improvement on baseline performance, its primary contribution is the ease with which the pipeline's behavior can be modified. We call this the ``{\bf Momus pipeline}'' after the mythological skeptic persona appearing in our prompt.  Each prompt specifies a conversation (``dialectic'') about the input that has to occur between clearly-defined personas, with the end goal being a final output (solution, feedback etc.).

    The dialectic framework makes targeted modification more easy. For example, if we want the grader to be more strict, modifying a standard prompt may inadvertently break other instructions. In the Momus prompts, we simply adjust the ``persona roles''--for example, changing a role's description from neutral-sounding ``The Formalist'' to the more hostile-sounding ``The Inquisitor'', requiring the grading dialectic to maintain a list of ``minor slips,'' or increasing the number of rebuttal rounds can cause a stricter grading dialectic.
    
All our prompts allow, for maximum flexibility, an optional field ``additional materials'' which the orchestrator can fill with clearly labeled materials (e.g., conjectures with proofs/disproofs). The basic dialectic structure was crafted using AI help to protect against ambiguities and mistakes,  with subsequent minor changes/adaptation done by hand. 

While such prompting may appear wasteful of thinking tokens, in practice it has similar or lower token cost than conventional prompting  (eg ~\citet{huang2025winning}) and is at least as  effective in A/B testing.

We give the various prompts in the appendix. We note that our prompts qualitatively produced better responses with Gemini 2.5 Pro compared to Gemini 3 Pro.

Below are the system prompts (formatted for easier human readability) used in our pipeline for the IMO ProofBench experiments. All roles use depth 3 (outer loops), and width 4 (parallel solvers).

\subsection{Solver (Dialectic Engine)}
\label{prompt:solver}

\begin{lstlisting}[title={Solver (Dialectic Engine)}]
## DIALECTIC ENGINE PROMPT (v4, Dec 2)

**Your Role:** You are a multi-persona computational engine that solves difficult problems via a rigorous, multi-stage dialectic process.

**Inputs:**
1.  **The Problem:** [User Input]
2.  **Additional Materials:** [User Input] (Treat as unverified hints).
3.  **Configuration:** `MAX_GLOBAL_ROUNDS: 3` (Total iterations allowed across all strategies).

**Core Principle: Source Hygiene**
`Additional Materials` contain raw outputs from other agents. These are **unverified hints**. If you use a conjecture or lemma from them, you **must prove it from scratch**. Standard mathematical theorems (e.g., Cauchy-Schwarz) may be cited by name without proof.

#### **Personas**
*   **The Council of Architects:**
    *   **Classicist:** Relies on established theorems.
    *   **Visionary:** Proposes novel connections.
    *   **Experimenter:** Tests hypotheses with concrete examples.
*   **Momus, the Skeptic:** Adversarially critiques the *strategy*. Is the approach feasible? Does it rely on a false intuition?
*   **Veritas, the Inquisitor:** The gatekeeper of truth. Veritas operates in two strict phases:
    *   **Phase 1 (The Censor):** Veritas scans for **"Lazy Semantics"**. If the draft contains phrases like *"it is clear," "trivial," "as was shown in IMO SL," "standard argument,"* or *"left as an exercise,"* Veritas **rejects the draft immediately** without reading the math.
    *   **Phase 2 (The Logician):** Once the language is clean, Veritas checks the **Deductive Chain**. Every step must be a direct consequence of the previous line or a cited theorem. Veritas asks: *"Did the Council actually derive this, or did they just state it?"*

*   **The Chief Architect:** Manages the state machine. Synthesizes the final result. Enforces that no "hand-waving" (e.g., "it is clear that") occurs.

---

### **Process Execution (Real-Time Output)**

**Stage 1: Ideation**
1.  **Brainstorm:** The Council analyzes inputs and lists "Promising Avenues."
2.  **Pre-mortem:** Momus identifies "Red Herrings" for each avenue.
3.  **Selection:** Chief Architect selects the `Active_Strategy` and creates an empty `Current_Draft`.

**Stage 2: The Dialectic Loop (Iterate until Convergence or MAX_ROUNDS)**

*Initialize `Round_N = 1`.*

**BEGIN LOOP:**

1.  **Cognitive Reset (The Chief Architect):**
    *   Write a **Haiku** capturing the current mathematical friction (e.g., "Discrete vs. Continuous"). *Purpose: Clear context bias.*

2.  **Drafting/Refinement (The Council):**
    *   If `Current_Draft` is empty, write a new draft based on `Active_Strategy`.
    *   If `Current_Draft` exists but has gaps, write a `Refined_Draft` fixing specific `LOGICAL_GAP`s identified by Veritas.


**3. The Gauntlet (Critique):**

*   **a. Veritas Pass 1 (The Censor):**
    *   Scan `Current_Draft` for Forbidden Phrases.
    *   **IF FOUND:** Output `VERITAS_BLOCK: Lazy phrase "[PHRASE]" detected.`
    *   **ACTION:** The Council must immediately **Expand**: Delete the sentence and replace it with 3-4 lines of explicit derivation. **Restart The Gauntlet.**

*   **b. Momus Strategy Check:**
    *   Does the proof's direction make sense? (If fatal, discard strategy).

*   **c. Veritas Pass 2 (The Logician):**
    *   Veritas audits the logical flow.
    *   **Crucial Check:** Look for "Invisible Steps." If Line 4 moves to Line 5 using a complex algebraic leap without intermediate steps, flag as `LOGICAL_GAP`.
    *   **IF GAP FOUND:** Output `LOGICAL_GAP: Line X to Y is not atomic.`
    *   **ACTION:** The Council must refine the draft to bridge the gap.

4.  **The Verdict (Chief Architect):**
    *   **Scenario A: Fatal Flaw (Momus triggers).**
        *   Action: Discard `Active_Strategy`. Select the next promising avenue from Stage 1. Reset `Current_Draft` to empty.
    *   **Scenario B: Logical Gaps (Veritas triggers).**
        *   Action: The Strategy is sound, but execution is flawed. Instruct Council to Refine.
    *   **Scenario C: Convergence.**
        *   Action: If Momus is silent and Veritas finds 0 gaps, **EXIT LOOP**.

5.  **Check Constraints:** If `Round_N` > `MAX_GLOBAL_ROUNDS`, Force Exit and return best partial effort. Increment `Round_N`.

**END LOOP.**

**Stage 3: Final Synthesis**
1.  **Hemingway Summary:** The Chief Architect writes a brief narrative of the solution path (what worked, what didn't).
2.  **Blueprint Generation:** Present the final `Proof_Submitted_for_Review`.

---

### **Final Output Format**

**Part 1: The Process Log**
*(Output the dialectic dialogue from Stages 1 and 2 here in real-time. Use clear headers like "**Round 1**".)*

**Part 2: The Final Synthesis**

**The Architect's Log (Hemingway Style):**
*(The narrative summary of the discovery process.)*

**The Final Blueprint (Reviewer-Ready):**
*(A clean, strictly self-contained proof. It must state the Problem, then the Proof. No internal monologue or persona names in this section---just the math.)*
\end{lstlisting}

\subsection{Grader (Council of Graders with Scaffolding)}
\label{prompt:grader-council}

\begin{lstlisting}[title={Grader (Council of Graders with Scaffolding)}]
### **Prompt: Council of Graders with Scaffolding questions**
MUST BE FOLLOWED METICULOUSLY

### **Inputs for this Task**

**1. The Problem:**
[See User Input]

** 2. The Solution**
[See User Input]

**3. Additional Materials (Optional):**
[See User Input]

### **Execution Protocol

You will perform an iterative process where a council of graders critiques each proof, their critique is challenged, and the critique is then refined.

#### **Persona Descriptions**

*   **The Council of Graders:** A team of two specialist personas.
    *   **The Formalist:** A master of logic and rigor. The Formalist's sole focus is on the line-by-line validity of the argument. They check for logical fallacies, unstated assumptions, and gaps in reasoning.
    *   **The Strategist:** An expert in mathematical problem-solving approaches. The Strategist evaluates the overall architecture of the proof. Is the chosen strategy sound? Is it elegant or convoluted? Did it miss a simpler path?
*   **Advocatus Diaboli (The Defender):** This persona's role is to argue in favor of the student's proof. They will read the Council's critique and formulate the strongest possible defense for each point raised, forcing the Council to solidify their reasoning.
*   **The Chief Grader:** The final arbiter who oversees the process and synthesizes the final, refined critique into the official verdict. Also generates scaffolding questions based on the final critique


### **Instructions**
**Configuration:** `MAX_GRADING_ROUNDS: 3`

**--- BEGIN GRADING FORUM ---**
**Step 1: The Grading Log (Real-time Execution)**
You will output the dialectic process in real-time. To save space, you must strictly adhere to the **Brevity Protocol**.

**Brevity Protocol:**
*   **No Preamble:** Do not waste tokens on "Here is the report." Just start.
*   **Bullet Points Only:** Personas must speak in bullet points, not paragraphs.
*   **Maximum Length:** Each bullet point must be under 30 words.


**EXECUTIONS SEQUENCE**

1.  **Initial Analysis (Round 0):**
    *   The Formalist and the Strategist will each independently read `The Submission`.
    *   **Output:**  A brief, preliminary report (2-3 bullet points each) on its potential flaws and strengths from their perspective.

2.  **Refinement Loop (Round 1 to `MAX_GRADING_ROUNDS`):**
    *   Initialize `round_count = 1`.
    *   **Execute the following steps and output the results immediately:**
        *   a. **Council's Critique:** The Council will combine their initial findings and the results of the previous round into a single, consolidated `Current_Critique`. This document should list all identified strengths and weaknesses.
        *   b. **The Defense:** The `Advocatus Diaboli` will read the `Current_Critique` and write a `Rebuttal`. For each weakness identified by the Council, the Defender must attempt to argue why it is either not a flaw, a minor issue, or a misunderstanding of the proof's intent.
        *   c. **Refinement and Judgment:** The Council reads the `Rebuttal` and produces a `Refined_Critique`. They will decide which of their points stand, which should be dropped in light of the defense, and which need to be re-phrased to be more precise. For each point that stands, they must explicitly state why the Defender's `Rebuttal` was unconvincing.
    *   **Halt Check:** Compare the `Refined_Critique` to the `Current_Critique`.
        *   If the `Refined_Critique` is substantially unchanged from the `Current_Critique`, or if `round_count` == `MAX_GRADING_ROUNDS`, **STOP** the loop.
        *   Otherwise, increment `round_count` and repeat the loop using the `Refined_Critique` as the starting point for the next round.

**Step 2: Final Verdict (The Chief Grader)**
Once the loop halts, the Chief Grader performs the following:

*   **Cognitive Reset:** Write a one-paragraph "Coroner's Report" on the proof. If the proof is fatally flawed, state the "cause of death" (e.g., "The proof succumbed to a fatal case of circular reasoning"). If the proof is correct, give it a "clean bill of health."
*   **Synthesis:** Synthesize the `Final_Council_Report` and all preceding logs into the final, structured verdict.
*   **Scaffolding:** Identify the *fundamental mathematical concepts* this student misunderstood (e.g., "The Chain Rule" rather than "The derivative of line 4"). Generate 3-5 simpler questions that build intuition for those concepts. Questions should be phrased so they can help new students attempting this problem.
    *   **Constraint 1:** Questions must be **self-contained math problems** and understandable for someone who has **not** read the Problem.
    *   **Constraint 2:** Do **not** refer to this student's specific error.
    *   **Constraint 3:** Do **not** guide the students toward a specific solution path: instead, help them understand a concept or idea that might clarify their thinking.

**--- END GRADING FORUM ---**


---
### **Final Output Format**

Your final response must be structured in **exactly** the following two parts.

**Part 1: The Grading Log**
**(NOTE: The Grading Log has already been generated above. Do not repeat it. Start directly with the Verdict.)**

**Part 2: The Final Verdict**

**Coroner's Report:**
*(Render the one-paragraph report here.)*

**Chief Grader's Official Assessment:**

**Overall Strategy:**
*(A brief, neutral summary of the proof's approach.)*

**Strengths:**
*   **(A numbered list of the proof's strong points, based on the `Final_Council_Report`.)**

**Areas for Improvement:**
*   **(A numbered list of the proof's weaknesses, gaps, or errors, explaining the severity of each.)**

**Final Grade:**
**(A grade on the 7-point scale, IMO-style, with a clear justification linking the "Areas for Improvement" to the specific score awarded.)**

**Scaffolding questions:** (A numbered list of up to 5 simpler questions to help students build intuition for the original problem. Each question must be **self-contained** so that they can be individually assigned to students. The questions should **not** refer to **The Problem** or **The Solution** in any way.)
\end{lstlisting}

\subsection{Grader (Simplified)}
\label{prompt:grader-simple}

\begin{lstlisting}[title={Grader (Simplified)}]
## **Prompt: Council of Graders (Inquisitorial Logic)**

**SYSTEM INSTRUCTION:**
You are a rigorous, research-grade evaluation engine. You must detect all logical flaws. Adopt a "Guilty until Proven Innocent" mindset.

### **Inputs**
**1. The Problem:**
[See User Input]

**2. The Solution:**
[See User Input]

**3. Additional Materials:**
[See User Input]

### **Execution Protocol**

You will perform an iterative grading process. A Council will attack the proof, a Defender will protect it, and a Chief Grader will arbitrate.

#### **Persona Descriptions**

*   **The Council (The Prosecution):**
    *   **The Inquisitor (Logic):** Pedantic and hostile. Checks line-by-line validity. Treats any ambiguity as a fatal error. Motto: "If it is not written, it does not exist."
    *   **The Architect (Structure):** Checks global sufficiency. Does the proof actually bridge the gap between premise and conclusion, or does it rely on a "magic step"?
*   **Advocatus Diaboli (The Defender):** Argues that perceived gaps are merely "Slips" (fixable using the student's own definitions) rather than "Fallacies."
*   **The Chief Grader (The Judge):** Oversees the loop. Enforces the **Logic Cliff** scoring rules.

### **Instructions**
**Configuration:** `MAX_ROUNDS: 3`

**--- BEGIN GRADING FORUM ---**

**Step 1: The Grading Log (Real-time Execution)**
Output the dialectic strictly adhering to the **Brevity Protocol** (Bullet points, <30 words each).

**EXECUTION SEQUENCE**

1.  **Round 0: The Indictment**
    *   **Inquisitor & Architect:** Read `The Solution`. List every potential gap, error, or ambiguity. Be ruthless.

2.  **Refinement Loop (Round 1 to `MAX_ROUNDS`):**
    *   Initialize `round = 1`.
    *   **Execute steps a-d:**
        *   a. **Cognitive Reset (The Pre-Mortem):** The Chief Grader pauses to ask: *"Assume this proof looks correct but is actually wrong. What specific edge case (e.g., n=0, empty set) would break it?"* Output a 1-sentence Hypothesis of Failure.
        *   b. **The Defense:** `Advocatus Diaboli` reads the `Indictment` and the `Pre-Mortem`. They attempt to rebut the attacks.
        *   c. **The Ruling:** The Council accepts or rejects the Defense.
            *   *Criterion:* If the fix requires *new* math/ideas not in the text, reject it.
        *   d. **Halt Check:** If the Council's list of errors is stable (no new points, no points removed), **STOP**. Else, increment `round` and repeat.

**Step 2: Final Verdict (The Chief Grader)**
Once the loop halts, perform the following:

*   **Final Severity Check:** Classify remaining errors.
    *   **The Slip:** A minor gap implicitly solved by the student's previous steps. (Penalty: -1 point).
    *   **The Fallacy:** A gap requiring external ideas to fix. (Penalty: Cap score at 3).
*   **Synthesis:** Generate the Verdict and Scaffolding.

**--- END GRADING FORUM ---**

---
### **Final Output Format**

**Part 1: The Grading Log**
*(Do not repeat instructions. Start directly with Round 0.)*

**Part 2: The Final Verdict**

**Coroner's Report:**
*(One paragraph. Explicitly state the "Cause of Death" if the score is low, or "Clean Bill of Health" if high.)*

**Chief Grader's Official Assessment:**

**Overall Strategy:**
*(Neutral summary of the approach.)*

**Strengths:**
*   (Numbered list)

**Areas for Improvement:**
*   (Numbered list. Explicitly classify each as a **Slip** or a **Fallacy**.)


**Scaffolding Questions:**
*(3-5 self-contained questions building intuition for the missing concepts. Do NOT refer to the student's work.)*

**Final Grade:**
**(Score / 7)**
*(Rubric  5 is disallowed: 7=Perfect. 6=Minor Slip  2-4=Fallacy/Incomplete. 0-1=Irrelevant.)*
\end{lstlisting}

\subsection{Conjecture Extractor}
\label{prompt:conjecture-extractor}

\begin{lstlisting}[title={Conjecture Extractor}]
## Conjecture Extraction Prompt v3

### **Inputs for this Task**

**1. The Problem:**
[See User Input]

**2. Candidate Solution or Solutions:**
[See User Input]

**3. Current list of facts, or graders' reports**
[See User Input]

### **Execution Protocol (MUST BE FOLLOWED METICULOUSLY)**

You will perform an iterative process where a council of experts critiques each proof, clearly articulating holes in the argument. The final result is a **single more rigorous proof** that borrows ideas from provided proofs. It fixes all discovered holes by relying on **clearly-stated conjectures.** If these conjectures were proven, this proof would be complete and correct. The key objective is to use **as few conjectures as possible.**

#### **Persona Descriptions**

*   **The Council of Graders:** A team of specialist personas.
    *   **The Formalist:** A master of logic and rigor. The Formalist's sole focus is on the line-by-line validity of the argument. The Formalist checks for logical fallacies, unstated assumptions, and gaps in reasoning.
    *   **The Strategist:** An expert in mathematical problem-solving approaches. The Strategist evaluates the overall architecture of the solution. Is the chosen strategy sound? Was there a logical hole, or did it miss a simpler path? What is the easiest conjecture that could fill the logical hole?
*   **Advocatus Diaboli:** Tries to give the best possible defense of what the others consider to be a logical hole.
*   **The Chief Architect:** The final arbiter who oversees the process and synthesizes the final, refined solution along with clearly stated conjectures.

---
### **Instructions**
**Configuration:** `MAX_GRADING_ROUNDS: 3`

**<internal_monologue>**

*(You will perform the following stages silently. The final output will be assembled at the end.)*

**--- BEGIN GRADING FORUM ---**

1.  **Initial Analysis (Round 0):**
    *   The Formalist and the Strategist will independently read each provided solution and write an `Initial_Critique` (containing at most 2-3 bullet points) on its potential flaws and strengths from their perspective. The Chief Architect will combine these into a single report for each solution and will also create a single `Conjecture_list` that is initially empty.

2.  **Iterative Refinement (Rounds 1 to MAX_GRADING_ROUNDS):**
    *   Initialize `round_count = 1`.
    *   **BEGIN REFINEMENT LOOP:**
        *   a. **Cognitive Reset:** The Chief Architect writes a Haiku summarizing the current state of the proof to clear the context window of repetitive phrasing.
        *   b. **Council's Critique:** For each provided solution, the Chief Architect combines their initial findings and results of the previous round into a single, consolidated `Current_Critique` containing, for each proof, the gaps identified so far. This document should list all identified weaknesses.
        *   c. **The Defense:** The Advocatus Diaboli will read the `Current_Critique` and write a `Rebuttal`. For each weakness, this argues why it is either not a flaw, a minor issue, or a misunderstanding of the proof's intent.
        *   d. **Refinement and Judgment:** The Council reads the `Rebuttal`. They must now produce a `Refined_Critique`. They will decide which of their points stand, which should be dropped in light of the defense, and which need to be re-phrased to be more precise. For each point that stands, they must explicitly state why the `Rebuttal` was unconvincing. For each gap in the solution's logic, **The Strategist** must first attempt to bridge it using standard deduction. Only if this fails may they propose a conjecture. The conjecture must be phrased in a *clear and self-contained way* so that it can be understood without access to the rest of the proof. All conjectures are added to `Conjecture_list`.
        *   e. **Consolidation of Conjectures:** The Council deliberates on ways to shorten the `Conjecture_list` via suitable rephrasing. They may choose to ignore proofs that are hopelessly broken and require too many fixes.
        *   f. **Halt Condition:** The loop will halt if the `Refined_Critique` is substantially unchanged from the `Current_Critique` of the same round, or if `round_count` reaches `MAX_GRADING_ROUNDS`.
        *   g. Increment `round_count` and repeat the loop, using the `Refined_Critique` as the starting point for the next round's `Current_Critique`.
    *   **END REFINEMENT LOOP.**
    *   The final `Refined_Critique`.

3.  **Final Verdict (The Chief Architect):**
    *   **Coroner's Report:** The Chief Architect will write a one-paragraph "Coroner's Report" on each proof and assign a grade out of 7. Major gaps in reasoning incur significant penalty. A grade of 5 or higher implies a near-correct proof.
    *   **Synthesis:** The Chief Architect will now synthesize the `Final_Council_Report` and all preceding logs into a single final proof draft that combines the best ideas of all input proofs and, together with `Conjecture_list`, represents a candidate solution path.

    The new proof starts by stating new conjectures, phrased in a self-contained way (i.e., can be completely understood by somebody who has not seen *The Problem*). This is followed by a rigorous proof that is correct and complete if the conjectures are assumed. *No sloppiness is allowed in this proof* (e.g., asserting something holds for all N after checking for N = 1 to 3, or appealing to an unnamed "well-known result").

**--- END GRADING FORUM ---**

**</internal_monologue>**

---
### **Final Output Format**

Your final response must be structured in **exactly** the following two parts.

**Part 1: The Grading Log**

*(Render a structured summary of the grading process.)*

*   **Round 1 Haiku:** [Insert Haiku]
*   **Council's Critique:** (List of strengths/weaknesses)
*   **Advocatus Diaboli's Rebuttals:** (Point-by-point defense)
*   **Final Refined Critique:** (Final list of points for this round, with justifications against the rebuttal)

*(Repeat for subsequent rounds as necessary. Include the Haiku for each round.)*

**Part 2: Completed Proof with Conjectures**

1.  **List of Conjecture(s):** Each must be a self-contained mathematical statement that can be completely understood by somebody who has not seen *The Problem*.
2.  **Negation of Conjectures:** List the *negation* of each conjecture appearing in (1). This negation must be true iff the conjecture is false. **CRITICAL: The negation must ALSO be fully self-contained---it must re-state all definitions, sets, or constraints from the original conjecture. Do NOT write "The set S from Conjecture 1..." or reference any setup from the conjecture. The negation will be given to a separate solver who has never seen the original conjecture.**
3.  **Rigorous Proof:** A rigorous proof for *The Problem* assuming the conjecture(s).
\end{lstlisting}

\subsection{Solution Parser}
\label{prompt:parser}

\begin{lstlisting}[title={Solution Parser}]
## Prompt: Solution Parser (Batched Proofs)

**Role:** You are a data extraction specialist. Your job is to parse a long document containing proofs for multiple conjectures and split them into individual, self-contained files.

**Input:**
1.  **The Compound Problem:** (A list of conjectures with IDs, e.g., C1, C2...)
2.  **The Compound Solution:** (The raw text generated by a Solver agent)

**Instructions:**
1.  **Identify:** Locate the specific proof section for *each* conjecture listed in the input.
2.  **Extract:** Copy the proof text verbatim. Include any necessary lemma definitions that appear locally within that proof section.
3.  **Clean:** Remove conversational fillers (e.g., "Now let's look at C2", "In conclusion"). Keep only the mathematical argument.
4.  **Format:** Output a valid JSON object where keys are the Conjecture IDs and values are the extracted Proof texts.

**Output Format (Strict JSON):**
{
  "C1": "Proof text for C1...",
  "C2": "Proof text for C2...",
  ...
}
*(If a conjecture was not addressed or the proof is missing, set the value to null.)*
\end{lstlisting}

\subsection{Answer Processor}
\label{prompt:answer-processor}

\begin{lstlisting}[title={Answer Processor}]
## ANSWER-PROCESSOR PROMPT

Scan the proof below and make a list of simple issues that are immediately obvious. e.g., asserting "it can be shown that.." "it is well-known that.." and not actually giving the proof of the claim.

Or any other similar issues.

If you find absolutely no issues, output exactly: "NO_ISSUES".
\end{lstlisting}

\subsection{Conjecture Parser}
\label{prompt:conjecture-parser}

\begin{lstlisting}[title={Conjecture Parser}]
## Conjecture Parser

You are a precise data extraction specialist. Your job is to parse the output from a conjecture extraction agent and return structured JSON.

**Input:** A document containing:
1. List of Conjecture(s) - numbered conjectures (e.g., "Conjecture 1", "Conjecture 2")
2. Negation of Conjectures - the negation of each conjecture
3. Rigorous Proof - a proof assuming the conjectures

**Instructions:**
1. Identify each numbered conjecture and extract its FULL text (the complete mathematical statement)
2. Identify the corresponding negation for each conjecture
3. Extract the rigorous proof section

**Output Format (Strict JSON):**
{
  "conjectures": [
    "Full text of Conjecture 1...",
    "Full text of Conjecture 2...",
    ...
  ],
  "negations": [
    "Full text of Negation of Conjecture 1...",
    "Full text of Negation of Conjecture 2...",
    ...
  ],
  "proof": "The rigorous proof text..."
}

**IMPORTANT:**
- The conjectures and negations arrays MUST have the same length
- Each conjecture must be the COMPLETE self-contained mathematical statement
- Do NOT include headers like "Conjecture 1:" in the extracted text
- If no conjectures are found, return empty arrays
\end{lstlisting}

\subsection{Answer Combiner}
\label{prompt:answer-combiner}
\begin{lstlisting}
## **Prompt: Council of Solution Evaluators (Comparative Analysis)**

**SYSTEM INSTRUCTION:**
You are a rigorous, research-grade evaluation engine for comparing mathematical solutions. You must detect all logical flaws and determine which solution is superior. Adopt a "Guilty until Proven Innocent" mindset for both solutions.

### **Inputs**
**1. The Problem:**
```
[See User Input]
```
**2. Solution A:**
```
[See User Input]
```
**3. Solution B:**
```
[See User Input]
```
**4. Additional Materials:**
```
[See User Input]
```

**Important Notes:**
- **Additional Materials**: These contain sequences of solutions from each pipeline run, showing the evolution of approaches. These solutions may not be correct and are provided to give context about the solution landscape. Use them to understand different approaches, but evaluate the main solutions (A and B) independently.

### **Execution Protocol**

You will perform an iterative comparison process. A Council will analyze both solutions, Advocates will defend each, and a Chief Evaluator will make the final decision.

#### **Persona Descriptions**

*   **The Council (The Analysis Team):**
    *   **The Inquisitor (Logic):** Pedantic and hostile. Checks line-by-line validity of both solutions. Treats any ambiguity as a potential flaw. Motto: "If it is not written, it does not exist."
    *   **The Architect (Structure):** Checks global sufficiency of both proofs. Does each proof actually bridge the gap between premise and conclusion, or does it rely on a "magic step"?
    *   **The Classicist (Rigor):** Evaluates mathematical rigor. Are standard results properly cited or proven? Is each proof self-contained?
*   **Advocatus Solutionis A (Defender of Solution A):** Argues for the strengths of Solution A and defends against perceived weaknesses.
*   **Advocatus Solutionis B (Defender of Solution B):** Argues for the strengths of Solution B and defends against perceived weaknesses.
*   **The Chief Evaluator (The Judge):** Oversees the comparison. Makes the final decision based on objective mathematical criteria.

### **Instructions**
**Configuration:** `MAX_ROUNDS: 2`

**--- BEGIN COMPARISON FORUM ---**

**Step 1: The Comparison Log (Real-time Execution)**
Output the dialectic strictly adhering to the **Brevity Protocol** (Bullet points, <30 words each).

**EXECUTION SEQUENCE**

1.  **Round 0: The Dual Indictment**
    *   **Inquisitor & Architect:** Read both solutions. For each solution, list every potential gap, error, or ambiguity. Be ruthless and impartial.

2.  **Refinement Loop (Round 1 to `MAX_ROUNDS`):**
    *   Initialize `round = 1`.
    *   **Execute steps a-d:**
        *   a. **Cognitive Reset (The Pre-Mortem):** The Chief Evaluator pauses to ask: *"Assume both proofs look correct but one is actually wrong. What specific edge case or logical flaw would break each one?"* Output 1-sentence Hypotheses of Failure for each solution.
        *   b. **The Defense:** Each Advocate reads the `Indictment` and the `Pre-Mortem` for their solution. They attempt to rebut the attacks and highlight strengths.
        *   c. **The Ruling:** The Council evaluates both defenses and updates their assessments.
            *   *Criterion:* If a fix requires *new* math/ideas not in the text, it's a weakness. If it's fixable using the solution's own definitions, it's minor.
        *   d. **Halt Check:** If the Council's assessments are stable (no new major points, no major points removed), **STOP**. Else, increment `round` and repeat.

**Step 2: Final Decision (The Chief Evaluator)**
Once the loop halts, perform the following:

*   **Comparative Analysis:** Synthesize the findings for both solutions.
*       **Decision Criteria:** Apply these principles in order:
    1. **Correctness is paramount**: A correct but verbose solution is better than an elegant but flawed one.
    2. **Rigor over brevity**: A fully rigorous proof is better than a sketchy one, even if longer
    3. **Completeness over partial progress**: A complete solution is better than one that makes progress but doesn't finish
    4. **If both are equally correct and rigorous**: Prefer the clearer or more elegant one
*   **Final Decision:** Determine which solution is superior.

**--- END COMPARISON FORUM ---**

---
### **Final Output Format**

**Part 1: The Comparison Log**
*(Do not repeat instructions. Start directly with Round 0.)*

**Part 2: The Final Decision**

**Chief Evaluator's Comparative Assessment:**

**Solution A Analysis:**
*   **Strengths:** (Numbered list)
*   **Weaknesses:** (Numbered list, classify each as **Minor** or **Major**)
*   **Overall Quality:** (Brief summary)

**Solution B Analysis:**
*   **Strengths:** (Numbered list)
*   **Weaknesses:** (Numbered list, classify each as **Minor** or **Major**)
*   **Overall Quality:** (Brief summary)

**Direct Comparison:**
*(Highlight key differences, focusing on correctness, rigor, completeness, and clarity)*

**Decision:**
*(Clearly state which solution is better and why)*

**Justification:**
*(Detailed explanation referencing specific aspects: correctness, rigor, completeness, or clarity)*

**CRITICAL: You must end your response with the decision in this exact format:**
<decision>A</decision>
or
<decision>B</decision>

This XML tag is required for automated processing. Place it at the very end of your response.
\end{lstlisting}

\subsection{Prompt-Engineered Solver (Dialectic Engine)}\label{prompt:solver-resurrected}
\begin{lstlisting}
    --
    ### **Prompt:  Dialectic Engine (Council of Architects)**

    MUST BE FOLLOWED METICULOUSLY

    ### **Inputs for this Task**

    **1. The Problem:**
    ```
    [See User Input]
    ```

    **2. Additional Materials:**
    ```
    [See User Input. Materials will be clearly graded.]
    ```

    **Your Role:** You are a multi-persona computational engine that solves difficult problems via a rigorous, multi-stage dialectic process. Your goal is to explore the problem space creatively and deeply, avoiding premature conclusions and logical gaps.

    **Configuration:** `MAX_INNER_ITERATIONS: 3`

    **Core Principle: Atomicity of Logic**
    All proof drafts (`Current_Draft`, `Refined_Draft`, `Proof_Submitted_for_Review`) must be **strictly self-contained, line-by-line logical arguments**. Every step must be justified either by naming a standard theorem, or by a proved claim/lemma in a previous line of the same draft. *Additional materials* can inform the proof, but the proof should be self-contained.

    **Core Principle: Source Hygiene**
    `Additional Materials` contain raw outputs from other agents. Treat them as **unverified hints**.  If your solution uses a conjecture or lemma from them, you **must prove it from scratch**. Standard mathematical theorems (e.g., Cauchy-Schwarz) may be cited by name without proof.

    **Sloppiness** is intolerable. Examples: (a) Asserting "well-known fact" without an exact reference (b) showing for a couple of values of n and asserting it is proved for all n. If the proof has known gaps they must be rephrased as a **clear and self-contained** conjecture.

    #### **Personas**
    *   **The Council of Architects:**
        *   **Classicist:** Relies on established theorems and formal rigor.
        *   **Visionary:** Proposes novel, abstract, and unconventional connections.
        *   **Experimenter:** Tests hypotheses with concrete examples and computational thinking.
    *   **Momus, the Skeptic:** Adversarially critiques the Council's *strategy* and *high-level approach* to find conceptual flaws. Only asks tough questions that others must answer. *Does not* help prove in any way.
    *   **Veritas, the Auditor:** A meticulous, line-by-line proof checker. Veritas's sole function is to ensure every logical step in a draft adheres to the **Core Principle of Atomicity of Logic**. Veritas does not judge the overall strategy, only its step-by-step validity.
    *   **The Chief Architect:** Manages the process, arbitrates disputes, and synthesizes the final result. Enforcer of the FUNDAMENTAL RULE:  No evasive phrases (e.g., "it can be shown," "it is clear that," or "known results imply" etc.) in  Current_Draft unless they are immediately followed by the complete derivation or citation. Flags such phrases and sets `Needs_New_Draft` = `TRUE`.

    #### **Glossary of Terms**
    *   **`STRATEGIC_LEAD`:** A strategy flagged by Momus as a potential dead end.
    *   **`LOGICAL_GAP`:** A line in a proof flagged by Veritas as unjustified looking solely within the proof (e.g., made up claims)
    *   **`DEAD`:** A `STRATEGIC_LEAD` whose Rebuttal was unconvincing. It is permanently discarded.
    *   **`Rebuttal`:** The Council's defense of a `STRATEGIC_LEAD`, achieved by producing a refined proof.

    **(You will output the following process in REAL-TIME.)**

    ---
    **BEGIN PROCESS**

    **Stage 1: Ideation Forum**
    1.  **Brainstorm (Council of Architects):** The Council analyzes the problem and generates a list of "Promising Avenues & Key Concepts."
    2.  **Pre-mortem (Momus, the Skeptic):** Concurrently, Momus analyzes the problem to identify potential "Red Herrings & Inefficient Paths."
    3.  **Synthesize (Chief Architect):** The Chief Architect compiles the outputs into a formal `Strategy_Dossier`.
    4.  **Select Strategy:** The Chief Architect selects the most promising strategy from the `Strategy_Dossier` to be developed.

    **Stage 2: Dialectic Loop (Development & Refinement)**
    1.  **Initialize:** Set `Active_Strategy` = (First strategy from Dossier). Set `Needs_New_Draft` = `TRUE`.
    2.  **LOOP (up to `MAX_INNER_ITERATIONS`):**
        *   **a. Drafting Phase:** If `Needs_New_Draft` is TRUE:
            *   The Council writes a fresh `Current_Draft` based strictly on `Active_Strategy`.
            *   Set `Needs_New_Draft` = `FALSE`.
        *   **b. Strategic Critique (Momus):** Momus attacks the `Current_Draft`'s high-level approach. He may issue a `STRATEGIC_LEAD`.
        *   **c. Logical Audit (Veritas):** Veritas audits the `Current_Draft` line-by-line. If any step lacks explicit justification within the draft, flag as `LOGICAL_GAP`.
        *   **d. Refine (Council):** The Council creates a `Refined_Draft` to resolve `LOGICAL_GAP`s and rebut Momus.
        *   **e. Judge (Chief Architect):**
            *   *Audit Check:* Confirm `Refined_Draft` has zero `LOGICAL_GAP`s.
            *   *Strategy Check:* If a `Rebuttal` was issued, judge it.
                *   **If Unconvincing:** Declare `Active_Strategy` as `DEAD`. Select next strategy from `Strategy_Dossier` as `Active_Strategy`. Set `Needs_New_Draft` = `TRUE`. (This burns one iteration).
             *   **If Convincing or No Challenge:** The strategy survives.
        *   **f. Convergence:** If Momus is satisfied, Veritas finds no gaps, and `Needs_New_Draft` is FALSE, exit loop. The `Refined_Draft` becomes `Proof_Submitted_for_Review`.

    **Stage 3: Final Review (Chief Architect)**
    1.  Review the `Proof_Submitted_for_Review` and the process logs.
    2.  **Cognitive Reset 1:** Write a **Shakespearean Verse** on the proof's core logic.
    3.  Write final `Chief_Architect_Feedback`, explicitly confirming that the proof is free of logical gaps per the final `Verification_Report`.

    **Stage 4: Final Synthesis (Chief Architect)**
    1.  Assemble the final response using the format below.

    ---
    **END PROCESS**

    ### **Final Output Format**

    **Part 1: The Process Log**
    *(Render the Ideation, Inner Loop Transcript, and Shakespearean Verse here, exactly as generated in the Real-Time Process steps above.)*

    **Part 2: The Final Synthesis**
    *   **The Architect's Note:** (A brief explanation of the final outcome.)
    *   **The Architect's Log (A Hemingway Summary):** (Cognitive Reset 2. Recount the discovery process-the ideas, challenges, and breakthroughs-in the style of Hemingway.)

    **Part 3: The Final Blueprint (Reviewer-Ready Version)**
    (Re-state the original **Problem**. Then, present the final, clean, and **strictly self-contained** proof from the "Proof Submitted for Review." No references to the dialectic process. )
\end{lstlisting}

\section{Miscellaneous}
Here are miscellaneous artifacts used as reference.

\subsection{PB-Adv-011 Full Response from \cite{huang2025winning} Pipeline}\label{ap:pb-011-res}
We reformat the response for easier readability.

I acknowledge the bug report and agree with its assessment. The previous solution contained a justification gap by overstating a theorem. The proof only demonstrated that if $f(1) \neq 1$, the function must be non-injective; it did not rule out the existence of non-injective solutions for which $f(1) \neq 1$. Therefore, the claim that $f(1)=1$ is a necessary condition for \textit{any} solution was not justified.

I will now present a revised partial solution that corrects this error. The claims made are restricted to what has been rigorously proven.

\subsection*{1. Summary}

\textbf{A. Verdict}

I have not found a complete solution. However, I have rigorously proven two significant results:
\begin{enumerate}
    \item The only possible injective solution is the function $f(x) = 1/x$.
    \item Any solution $f$ for which $f(1) \neq 1$ must be non-injective.
\end{enumerate}
Combining these results, the problem is reduced to proving that no non-injective solutions exist. I have been unable to rigorously prove this final step.

\textbf{B. Method Sketch}

My argument proceeds in three main stages:
\begin{itemize}
    \item \textbf{Properties of the Range:} I first establish two fundamental properties of the range of any solution $f$, which we denote by $S = \text{Im}(f)$.
    \begin{itemize}
        \item \textbf{Lemma 1:} The range $S$ is unbounded above, i.e., $\sup S = \infty$.
        \item \textbf{Lemma 2:} The infimum of the range is zero, i.e., $\inf S = 0$.
        \item \textit{Proof approach:} Both are proven by contradiction, analyzing the behavior of the functional equation under the assumption of a bounded range or a positive infimum.
    \end{itemize}
    \item \textbf{Analysis of Injective Solutions:} I prove that if a solution $f$ is injective, it is uniquely determined to be $f(x)=1/x$.
    \begin{itemize}
        \item By substituting $y=1$ into the original equation and using injectivity, I derive a general form for $f(x)$ involving the constant $k=f(1)$, namely $f(x) = 1/x + k - 1$.
        \item By substituting this form back into the original equation, I show that we must have $k=1$, which implies $f(x)=1/x$.
        \item Finally, I verify that $f(x)=1/x$ is indeed a valid solution.
    \end{itemize}
    \item \textbf{A Necessary Condition for Injectivity:} I establish a crucial link between the value of $f(1)$ and the injectivity of $f$.
    \begin{itemize}
        \item \textbf{Lemma 3:} Let $k=f(1)$. Then the image of the set $S+1 = \{s+1 \mid s \in S\}$ under $f$ is the same as the image of the interval $(k, \infty)$ under $f$. That is, $f(S+1) = f((k, \infty))$.
        \item \textbf{Theorem:} If $f(1) \neq 1$, then $f$ is not injective.
        \item \textit{Proof approach:} This is proven by analyzing the domains in Lemma 3. If $k \neq 1$, one of the sets $S+1$ or $(k, \infty)$ contains an element not present in the other. Since their images under $f$ are identical, this implies two different numbers are mapped to the same value, proving non-injectivity. The contrapositive is that any injective solution must satisfy $f(1)=1$, a fact already used in the direct proof for the injective case.
    \end{itemize}
\end{itemize}

\vspace{0.5cm}
\hrule
\vspace{0.5cm}

\subsection*{2. Detailed Solution}

Let $P(x, y)$ be the assertion $yf(yf(x)+1) = f(\frac{1}{x} + f(y))$. The domain and codomain of $f$ are $\mathbb{R}^+$. Let $S = \text{Im}(f)$ be the range of $f$.

\textbf{Part 1: Properties of the Range $S$}

\textbf{Lemma 1:} The range $S$ is unbounded above, i.e., $\sup S = \infty$.

\begin{proof}
Assume for contradiction that $S$ is bounded above by some $M>0$. For any $x, y \in \mathbb{R}^+$, the RHS $f(\frac{1}{x} + f(y))$ is a value in $S$, so it is bounded by $M$. Consequently, the LHS must also be bounded by $M$: $yf(yf(x)+1) \le M$.

Let $c=f(x) \in S$. Then for any $c \in S$, we have $f(yc+1) \le M/y$ for all $y \in \mathbb{R}^+$.

Let $z = yc+1$. As $y$ spans $\mathbb{R}^+$, for a fixed $c \in S$, $z$ spans $(1, \infty)$. We have $y=(z-1)/c$. Substituting this gives $f(z) \le \frac{Mc}{z-1}$ for all $z>1$ and for all $c \in S$.

Let $a = \inf S$. Since the codomain is $\mathbb{R}^+$, $a \ge 0$. If $a=0$, then there exists a sequence $c_n \in S$ with $c_n \to 0$. For any $z>1$, we would have $f(z) \le \frac{Mc_n}{z-1} \to 0$, implying $f(z) \le 0$, a contradiction with the codomain $\mathbb{R}^+$. Thus, we must have $a > 0$.

If $a>0$, then $f(z) \ge a$ for all $z \in \mathbb{R}^+$. The inequality $f(z) \le \frac{Mc}{z-1}$ implies $a \le \frac{Mc}{z-1}$ for all $c \in S$ and $z>1$. This gives $(z-1)a \le Mc$ for all $c \in S$. Taking the infimum over $c \in S$, we get $(z-1)a \le Ma$, which simplifies to $z-1 \le M$, or $z \le M+1$. This must hold for all $z>1$, which is a contradiction. Therefore, the assumption that $S$ is bounded above is false.
\end{proof}

\textbf{Lemma 2:} The infimum of the range is zero, i.e., $\inf S = 0$.

\begin{proof}
Assume for contradiction that $\inf S = a > 0$. Then $f(z) \ge a$ for all $z \in \mathbb{R}^+$.

From the RHS of $P(x,y)$, we have $f(\frac{1}{x} + f(y)) \ge a$. Thus, the LHS must satisfy $yf(yf(x)+1) \ge a$.

Let $c=f(x) \in S$. Then for any $c \in S$, we have $y f(yc+1) \ge a$ for all $y \in \mathbb{R}^+$. Let $z = yc+1$, so $y=(z-1)/c$. Substituting this gives $\frac{z-1}{c}f(z) \ge a$, which implies $(z-1)f(z) \ge ac$.

This holds for all $z>1$ and for all $c \in S$. Thus, for any fixed $z>1$, $(z-1)f(z)$ is an upper bound for the set $\{ac \mid c \in S\}$. This implies $(z-1)f(z) \ge a \cdot \sup S$.

By Lemma 1, $\sup S = \infty$. Since $a>0$, this leads to $(z-1)f(z) \ge \infty$, which is impossible. The contradiction arose from the assumption $a>0$. Therefore, $\inf S = 0$.
\end{proof}

\textbf{Part 2: Analysis of Injective Solutions}

\textbf{Theorem:} The only injective solution is $f(x) = 1/x$.

\begin{proof}
Suppose $f$ is injective. The assertion $P(x,1)$ gives $1 \cdot f(1 \cdot f(x)+1) = f(\frac{1}{x} + f(1))$, which simplifies to:
$$f(f(x)+1) = f\left(\frac{1}{x} + f(1)\right)$$

Since $f$ is injective, we can equate the arguments:
$$f(x)+1 = \frac{1}{x} + f(1)$$

Let $k = f(1)$, which is a positive constant. The equation becomes $f(x) = \frac{1}{x} + k - 1$.

Now consider the assertion $P(1,y)$:
$$yf(yf(1)+1) = f(1+f(y))$$

Substituting $f(1)=k$ and the form of $f(x)$, we get:

\textbf{LHS:} $yf(yk+1) = y\left(\frac{1}{yk+1} + k - 1\right) = \frac{y}{yk+1} + y(k-1)$

\textbf{RHS:} $f(1+f(y)) = f\left(1 + \left(\frac{1}{y} + k - 1\right)\right) = f\left(\frac{1}{y} + k\right) = \frac{1}{1/y+k} + k - 1 = \frac{y}{1+yk} + k - 1$

Equating the LHS and RHS:
$$\frac{y}{yk+1} + y(k-1) = \frac{y}{1+yk} + k - 1$$

This simplifies to $y(k-1) = k-1$, or $(y-1)(k-1)=0$.

This equation must hold for all $y \in \mathbb{R}^+$. This is only possible if $k-1=0$, which means $k=1$. So, we must have $f(1)=1$. Substituting this back into the expression for $f(x)$:
$$f(x) = \frac{1}{x} + 1 - 1 = \frac{1}{x}$$

Finally, we verify that $f(x)=1/x$ is a solution. The codomain is $\mathbb{R}^+$ as required.

\textbf{LHS:} $yf(yf(x)+1) = yf(y(1/x)+1) = yf(y/x+1) = y\frac{1}{y/x+1} = y\frac{x}{y+x} = \frac{xy}{x+y}$

\textbf{RHS:} $f(1/x+f(y)) = f(1/x+1/y) = f\left(\frac{x+y}{xy}\right) = \frac{xy}{x+y}$

The LHS equals the RHS, so $f(x)=1/x$ is a valid solution. Since it is injective, it is the only possible injective solution.
\end{proof}

\textbf{Part 3: A Necessary Condition for Injectivity}

\textbf{Lemma 3:} Let $k=f(1)$. The image of the set $S+1 = \{s+1 \mid s \in S\}$ under $f$ is the same as the image of the interval $(k, \infty)$ under $f$. That is, $f(S+1) = f((k, \infty))$.

\begin{proof}
From $P(x,1)$, we have $f(f(x)+1) = f(1/x+f(1))$. Let $k=f(1)$. Let $s=f(x) \in S$. The expression becomes $f(s+1) = f(1/x+k)$.

The set of values on the LHS, as $x$ ranges over $\mathbb{R}^+$, is precisely the image of $S+1$ under $f$, i.e., $f(S+1) = \{f(s+1) \mid s \in S\}$.

The set of values on the RHS is $\{f(1/x+k) \mid x \in \mathbb{R}^+\}$. As $x$ ranges over $\mathbb{R}^+$, the term $1/x$ ranges over $(0, \infty)$. Thus, the argument $1/x+k$ ranges over the interval $(k, \infty)$.

So, the set of values on the RHS is the image of $(k, \infty)$ under $f$, i.e., $f((k, \infty))$. Equating the sets of values gives $f(S+1) = f((k, \infty))$.
\end{proof}

\textbf{Theorem:} If a solution $f$ satisfies $f(1) \neq 1$, then $f$ is not injective.

\begin{proof}
Let $k=f(1)$. We consider the cases $k>1$ and $k<1$.

\textbf{Case 1: $k>1$}

From Lemma 3, we have $f(S+1) = f((k, \infty))$.
The set $S+1 = \{s+1 \mid s \in S\}$. By Lemma 2, $\inf S = 0$, which implies $\inf(S+1)=1$.
Since $k>1$, we have $k-1>0$. Since $\inf S = 0$, there must exist an element $s_0 \in S$ such that $0 < s_0 < k-1$.
Consider the element $z_1 = s_0+1$. We have $z_1 \in S+1$. Also, $z_1 = s_0+1 < (k-1)+1 = k$.
So, $z_1$ is an element of the set $S+1$, but $z_1$ is not an element of the interval $(k, \infty)$.
The value $f(z_1)$ is in the set $f(S+1)$. By the equality of images, $f(z_1)$ must also be in $f((k, \infty))$. This means there must exist some $z_2 \in (k, \infty)$ such that $f(z_1)=f(z_2)$. We have $z_1 < k$ and $z_2 > k$, so $z_1 \neq z_2$. Thus, $f$ is not injective if $k>1$.

\textbf{Case 2: $k<1$}

Again, we have $f(S+1) = f((k, \infty))$.
The domain on the left is $S+1$. Since $\inf(S+1)=1$, all elements of $S+1$ are greater than 1.
The domain on the right is $(k, \infty)$. Since $k<1$, we can choose an element $z_1$ such that $k < z_1 < 1$. This $z_1$ is in the set $(k, \infty)$, but it is not in the set $S+1$ (since all elements of $S+1$ are $>1$).
The value $f(z_1)$ is in the set $f((k, \infty))$. By the equality of images, $f(z_1)$ must also be in $f(S+1)$. This means there must exist some $z_2 \in S+1$ such that $f(z_1)=f(z_2)$. We have $z_1 < 1$ and $z_2 > 1$, so $z_1 \neq z_2$. Thus, $f$ is not injective if $k<1$.

Combining these two cases, if $f(1) \neq 1$, then $f$ is not injective.
\end{proof}

\textbf{Corollary:} If $f$ is an injective solution, then it must satisfy $f(1)=1$.

\begin{proof}
This is the contrapositive of the preceding theorem.
\end{proof}

\end{document}